\pdfoutput=1
\documentclass{article}

\usepackage{microtype}
\usepackage{graphicx}
\usepackage{booktabs} 
\usepackage{enumitem}
\usepackage{fontawesome5}

\usepackage{subcaption}
\usepackage{hyperref}
\usepackage{multirow}
\usepackage{caption}
\usepackage{amsmath}
\usepackage{amssymb}
\usepackage{mathtools}
\usepackage{amsthm}
\usepackage{algorithm}
\usepackage{algpseudocode}
\usepackage{url}

\usepackage[accepted]{icml2025}

\usepackage[capitalize,noabbrev]{cleveref}

\theoremstyle{plain}

\theoremstyle{definition}

\theoremstyle{remark}

\usepackage[textsize=tiny]{todonotes}

\usepackage{makecell}
\usepackage[table]{xcolor}
\usepackage[most]{tcolorbox}
\usepackage{pifont}
\usepackage{tabularx} 

\usepackage{tikz}
\newcommand*\emptycirc[1][1ex]{\raisebox{-0.38ex}{\tikz\draw (0,0) circle (#1);}}
\newcommand*\halfcirc[1][1ex]{%
	\begin{tikzpicture}
        \raisebox{-0.38ex}{
	\draw[fill] (0,0)-- (90:#1) arc (90:270:#1) -- cycle ;
	\draw (0,0) circle (#1);}
	\end{tikzpicture}}
\newcommand*\fullcirc[1][1ex]{\raisebox{-0.38ex}{\tikz\fill (0,0) circle (#1);}}

\definecolor{t1}{RGB}{198,226,255}
\definecolor{t2}{RGB}{255,214,233}
\definecolor{t3}{RGB}{200,247,220}
\definecolor{t4}{RGB}{255,247,174}
\definecolor{t5}{RGB}{255,227,179}
\definecolor{t6}{RGB}{229,209,255}
\definecolor{t7}{RGB}{185,243,255}
\definecolor{t8}{RGB}{231,231,231}
\definecolor{t9}{RGB}{209,240,239}
\definecolor{t10}{RGB}{246,231,194}
\definecolor{t11}{RGB}{218,245,196}

\definecolor{t1dark}{RGB}{68,116,165}    
\definecolor{t2dark}{RGB}{165,94,130}    
\definecolor{t3dark}{RGB}{70,157,90}     
\definecolor{t4dark}{RGB}{165,157,64}    
\definecolor{t5dark}{RGB}{165,117,64}    
\definecolor{t6dark}{RGB}{119,99,165}    
\definecolor{t7dark}{RGB}{55,133,165}    
\definecolor{t8dark}{RGB}{120,120,120}   
\definecolor{t9dark}{RGB}{79,110,109}    
\definecolor{t10dark}{RGB}{116,101,54}   
\definecolor{t11dark}{RGB}{88,135,96}    

\newtcolorbox{examplebox}{
  breakable,                 
  colback=gray!15,           
  colframe=gray!50,          
  boxrule=0.5pt,             
  left=3pt, right=3pt,       
  top=2pt, bottom=2pt,       
  boxsep=1pt,                
  arc=5pt                    
}

\newtcolorbox{guidelinebox}{
  breakable,
  enhanced,
  colback=white,            
  colframe=t1dark!80,       
  colbacktitle=t1dark!20,   
  coltitle=black,           
  fonttitle=\bfseries,      
  boxrule=0.6pt,            
  left=4pt, right=4pt,
  top=3pt, bottom=3pt,
  arc=5pt,                  
  attach boxed title to top left={
    yshift=-2mm,
    xshift=3mm
  },
  boxed title style={
    arc=3pt,
    boxrule=0pt,
    left=3pt, right=3pt, top=1pt, bottom=1pt
  }
}

\icmltitlerunning{LongBench Pro}

\begin{document}

\twocolumn[

\icmltitle{\raisebox{-0.4em}{\includegraphics[height=1.5em]{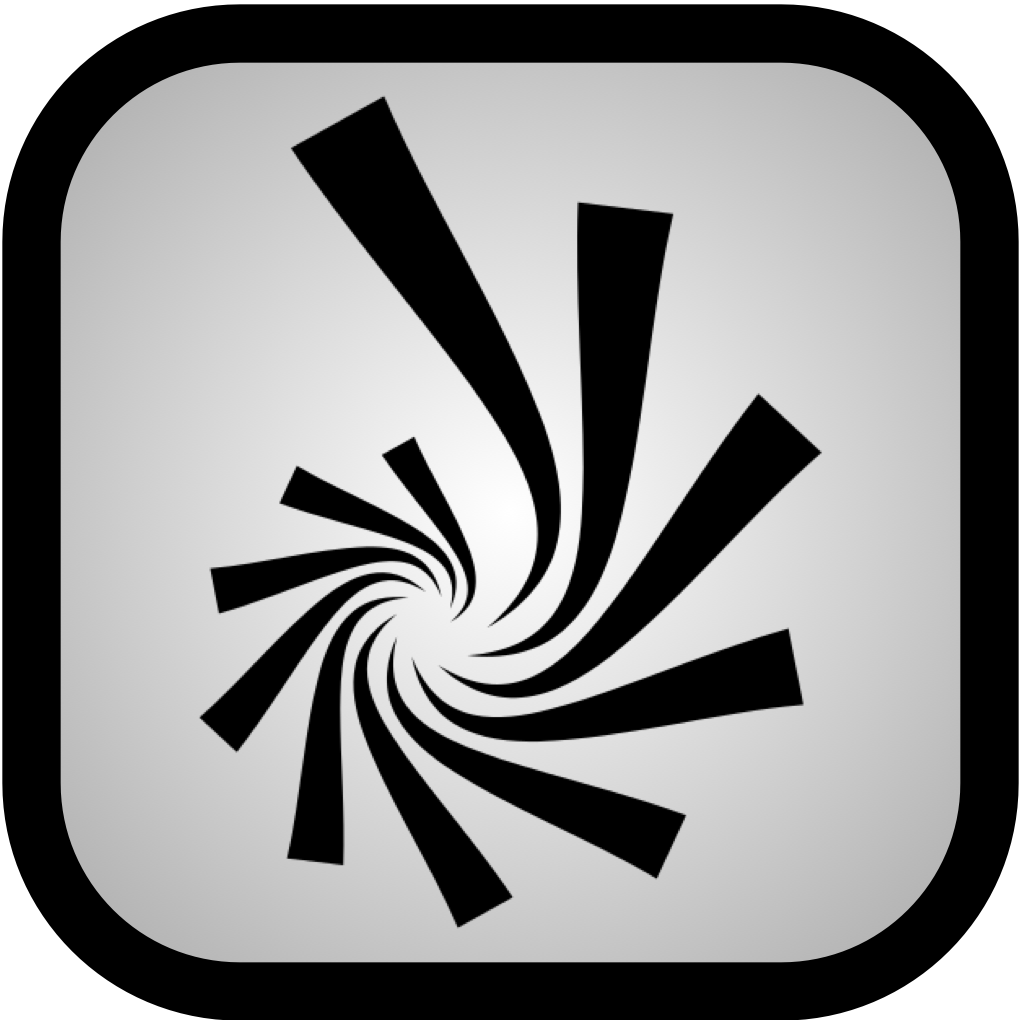}} LongBench Pro: A More Realistic and Comprehensive Bilingual Long-Context Evaluation Benchmark}

\icmlsetsymbol{equal}{*}

\begin{icmlauthorlist}
\icmlauthor{Ziyang Chen}{iie,ucas}
\icmlauthor{Xing Wu}{iie}
\icmlauthor{Junlong Jia}{buaa}
\icmlauthor{Chaochen Gao}{iie,ucas}
\icmlauthor{Qi Fu}{xhs}
\icmlauthor{Debing Zhang}{xhs}
\icmlauthor{Songlin Hu}{iie,ucas}

\end{icmlauthorlist}
\icmlaffiliation{buaa}{School of Artificial Intelligence, Beihang University}
\icmlaffiliation{iie}{Institute of Information Engineering, Chinese Academy of Sciences}
\icmlaffiliation{ucas}{School of Cyber Security, University of Chinese Academy of Sciences}
\icmlaffiliation{xhs}{Xiaohongshu Inc}

\icmlcorrespondingauthor{Xing Wu, Songlin Hu}{wuxing@iie.ac.cn, husonglin@iie.ac.cn}


\vskip 0.3in

\begin{center}
  \vspace{-0.15in}
  \resizebox{\textwidth}{!}{\includegraphics{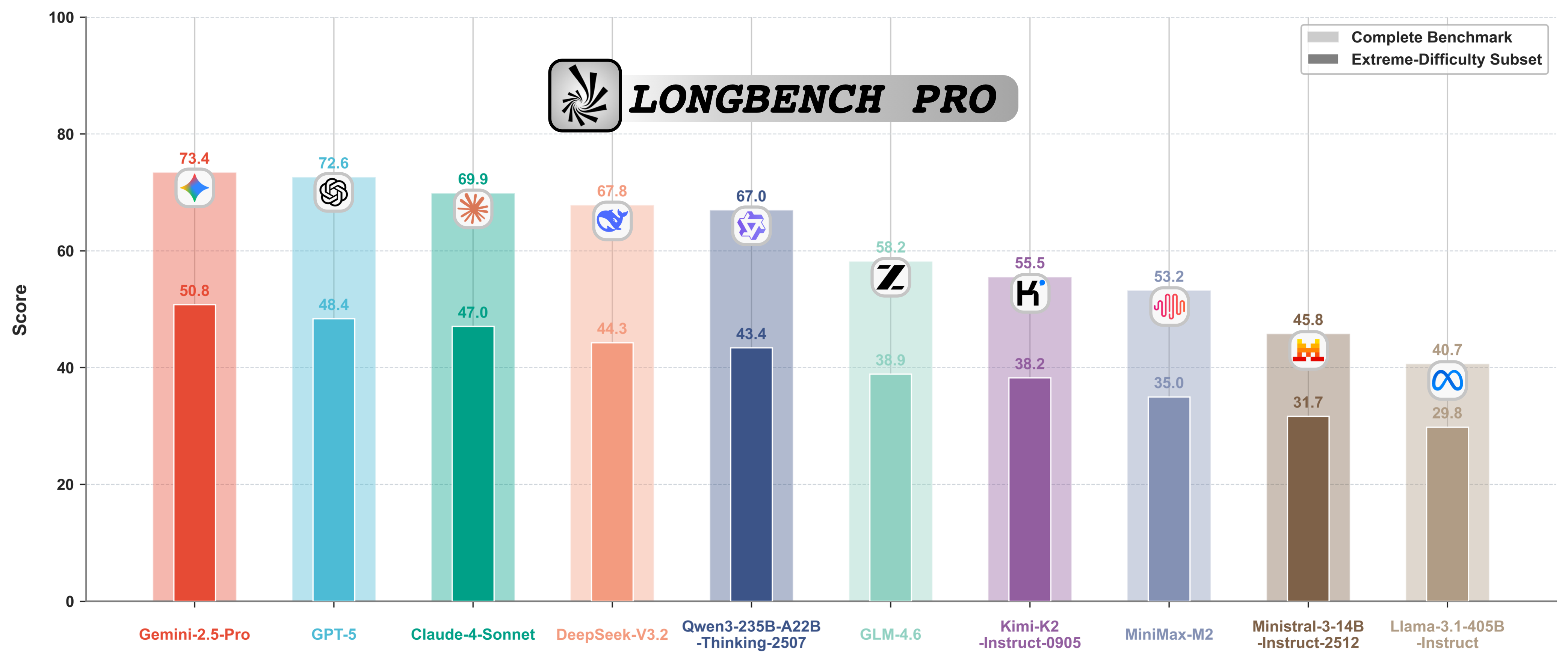}}
  \protect\captionof{figure}{Performance of advanced long-context LLMs on LongBench Pro.}
  \label{fig:contrast}
  \vspace{0.05in}
\end{center}

]



\printAffiliationsAndNotice{}  

\begin{abstract}
The rapid expansion of context length in large language models (LLMs) has outpaced existing evaluation benchmarks. Current long-context benchmarks often trade off scalability and realism: synthetic tasks underrepresent real-world complexity, while fully manual annotation is costly to scale to extreme lengths and diverse scenarios. We present \textbf{LongBench Pro}, a more realistic and comprehensive bilingual benchmark of 1,500 naturally occurring long-context samples in English and Chinese spanning 11 primary tasks and 25 secondary tasks, with input lengths from 8k to 256k tokens. LongBench Pro supports fine-grained analysis with task-specific metrics and a multi-dimensional taxonomy of \textit{context requirement} (full vs.\ partial dependency), \textit{length} (six levels), and \textit{difficulty} (four levels calibrated by model performance). To balance quality with scalability, we propose a \textit{Human-Model Collaborative Construction} pipeline: frontier LLMs draft challenging questions and reference answers, along with design rationales and solution processes, to reduce the cost of expert verification. Experts then rigorously validate correctness and refine problematic cases. Evaluating 46 widely used long-context LLMs on LongBench Pro yields three findings: (1) long-context optimization contributes more to long-context comprehension than parameter scaling; (2) effective context length is typically shorter than the claimed context length, with pronounced cross-lingual misalignment; and (3) the ``thinking'' paradigm helps primarily models trained with native reasoning, while mixed-thinking designs offer a promising Pareto trade-off. In summary, LongBench Pro provides a robust testbed for advancing long-context understanding.
\end{abstract}


\section{Introduction}

\begin{table*}[t!]
  \centering
  \resizebox{\textwidth}{!}{%
  \begin{tabular}{cccccccc}
  \toprule
  \multirow{2}{*}{\textbf{Benchmark}} & \multirow{2}{*}{\textbf{Text Type}} & \multirow{2}{*}{\textbf{\#Task}} & \multirow{2}{*}{\textbf{Metric}} & \multirow{2}{*}{\textbf{Language}} & \multicolumn{3}{c}{\textbf{Dimensional Categorization}} \\
  \cmidrule(l){6-8}
   &  &  &  &  & \textbf{Ctx-Req} & \textbf{Length} & \textbf{Difficulty} \\
  \midrule
  RULER~\cite{RULER} & Fully Synthetic & 4 & Single & EN & \emptycirc & \fullcirc & \emptycirc \\
  $ \infty $BENCH~\cite{inftyBench} & Synthetic, Natural & 6 & Diverse & EN, ZH & \emptycirc & \emptycirc & \emptycirc \\
  CLongEval~\cite{Clongeval} & Synthetic, Natural & 7 & Diverse & ZH & \fullcirc & \halfcirc & \emptycirc \\
  HELMET~\cite{Helmet} & Synthetic, Natural & 7 & Diverse & EN & \emptycirc & \fullcirc & \emptycirc \\
  LongBench v2~\cite{longbenchv2} & Fully Natural  & 6 & Single & EN & \emptycirc & \halfcirc & \halfcirc \\
  \midrule
  \rowcolor{gray!20} 
  \textbf{LongBench Pro (Ours)} & \textbf{Fully Natural} & \textbf{11} & \textbf{Diverse} & \textbf{EN, ZH} & \fullcirc & \fullcirc & \fullcirc \\
  
  \bottomrule
  \end{tabular}}
  \caption{\textbf{Comparison of long-context benchmarks.} Ctx-Req denotes the Context Requirement dimension. \scalebox{0.8}{\emptycirc}~indicates the absence of dimensional categorization; \scalebox{0.8}{\fullcirc}~indicates the presence of detailed categorization in this dimension  (Ctx-Req $\ge 2$; Length $> 3$; Difficulty $> 2$); \scalebox{0.8}{\halfcirc}~indicates the presence of categorization with a coarser granularity that does not meet the fine-grained criteria.}
  
  \label{tab:bench_comparison}
\end{table*}

\begin{figure*}[!h]
  \centering
  \includegraphics[width=\textwidth]{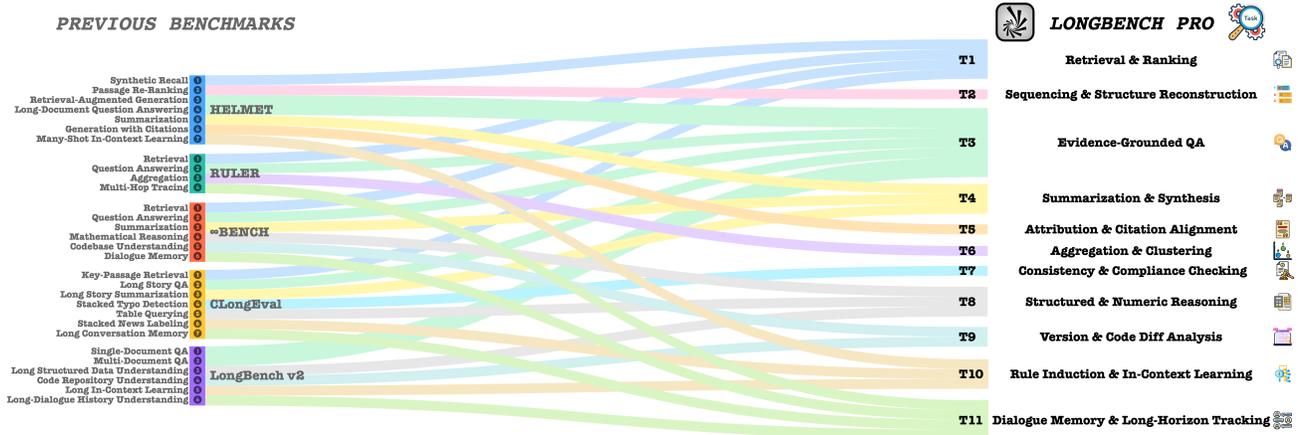}
  \caption{Task mapping between LongBench Pro and existing benchmarks.}
  \label{fig:task_map}
  \end{figure*}

Understanding and reasoning over long contexts has become a core capability of large language models (LLMs). With advances in architectures and computational resources, context length has continuously expanded~\cite{gemini25,gpt5,claude4}, making it possible for LLMs to tackle complex long-context tasks such as large-scale codebase analysis, legal document comprehension, and systematic literature review. This rapid progress demands that long-context benchmarks evolve swiftly to accurately assess models' true long-context capabilities and facilitate open-source models in closing the gap with closed-source models.

Several long-context evaluation benchmarks currently exist. Synthetic benchmarks~\cite{RULER,Helmet} enable controlled evaluation at scale, while manually annotated benchmarks~\cite{Clongeval,longbenchv2} ensure authenticity through complex tasks and realistic scenarios. Although these benchmarks have effectively guided model development over the past few years, they are increasingly insufficient (in terms of task coverage, difficulty level, etc.) for evaluating next-generation models, as context lengths extend to millions of tokens and model capabilities continue to advance rapidly. This creates a pressing need for new benchmarks that satisfy more authentic and comprehensive evaluation of diverse long-context capabilities.

To address this urgent need, we introduce \textbf{LongBench Pro}, a realistic and comprehensive long-context evaluation benchmark containing 1,500 bilingual samples in English and Chinese. Built entirely on authentic, naturally occurring long documents, LongBench Pro encompasses 11 primary tasks and 25 secondary tasks, covering the full spectrum of long-context capabilities assessed by all existing benchmarks, as shown in Figure~\ref{fig:task_map}. It employs diverse evaluation metrics for fine-grained measurement and introduces a multi-dimensional taxonomy: (1) \textit{Context Requirement}—\textit{Full} context (global integration) versus \textit{Partial} context (localized retrieval); (2) \textit{Length}—six levels uniformly distributed from 8k to 256k tokens to analyze scaling behavior; and (3) \textit{Difficulty}—four levels ranging from Easy to Extreme, calibrated based on model performance. Table~\ref{tab:bench_comparison} summarizes the key differences between LongBench Pro and existing benchmarks.

Creating such a benchmark requires moving beyond traditional construction approaches. Synthetic methods, despite their scalability, cannot fully capture the semantic complexity of realistic long context scenarios. Conversely, purely manual annotation becomes prohibitively expensive and cognitively demanding at extreme lengths, limiting practicality and scalability. To achieve both authenticity and scalability, we propose a novel \textit{Human-Model Collaborative Construction} strategy that synergizes the strengths of advanced LLMs and human expertise. In this strategy, models generate challenging questions and reference answers, along with design rationales and solution processes, from authentic, long documents corresponding to realistic, long-context scenarios. Meanwhile, human experts act as rigorous critics to verify correctness, filter flawed samples, and calibrate difficulty levels. This approach achieves high realism through the use of authentic documents, ensures quality through expert verification, and reduces costs through model involvement, enabling the efficient production of challenging, comprehensive, and high-quality evaluation samples.

We conduct extensive evaluations of 46 widely studied long-context LLMs on LongBench Pro and derive three main findings:

\noindent\textbf{(1) Long-context optimization surpasses parameter scaling for improving long-context comprehension.} Performance gains from extending effective context length significantly exceed those from scaling model size. This marks a shift from the traditional ``scale-first'' to a ``context-optimization-first'' paradigm.

\noindent\textbf{(2) Context length–capability gaps and language bias constrain performance consistency.} Models' effective context lengths often fall short of their claimed context lengths, and cross-lingual performance remains misaligned. While stronger models reduce these gaps, they persist across most long-context models.

\noindent\textbf{(3) The thinking paradigm requires native training to overcome long-context bottlenecks.} Thinking substantially improves long-context performance, but primarily in models trained with native reasoning. Conventional instruct models show limited or degraded gains when prompted to think. Mixed-thinking models combining fast response and deep reasoning achieve Pareto-optimal performance and may define the future paradigm.

Our contributions are threefold:
\begin{itemize}
    \item We release \textbf{LongBench Pro}, a realistic and comprehensive bilingual benchmark with 1,500 samples and multi-dimensional categorization for rigorous long-context evaluation.
    \item We validate a \textbf{Human-Model Collaborative Construction} strategy that transcends the cost-quality trade-off, enabling high-quality sample generation at significantly lower cost than purely manual annotation.
    \item We provide \textbf{comprehensive analysis} of 46 long-context LLMs, revealing insights into context length scaling, cross-lingual performance, and reasoning mechanisms.
\end{itemize}

\section{Task Framework of LongBench Pro}

To systematically characterize long-context capabilities, we organize LongBench Pro into a two-level task taxonomy. We consolidate task formulations from prior long-context benchmarks into 11 primary categories. Figure~\ref{fig:task_map} maps our taxonomy to existing benchmark tasks, illustrating that LongBench Pro covers all the core capability dimensions evaluated by existing benchmarks.

Beyond task type, we also introduce the \textit{context requirement}, an orthogonal dimension capturing how globally a model must depend on the input document:

\vspace{-5pt}

\begin{itemize}
    \item \textbf{Full}: solving the task requires integrating evidence dispersed across multiple, distant spans of the document, emphasizing integration and reasoning;
    \item \textbf{Partial}: solving the task primarily relies on localized spans, emphasizing localization and retrieval.
\end{itemize}

\vspace{-5pt}

Crossing the 11 primary categories with context requirements to design 25 secondary categories. Table~\ref{tab:task_definitions} lists the full taxonomy, and Appendix~\ref{appendix:task_definitions} provides detailed task definitions.

\begin{table*}[!t]
  \centering
  \resizebox{\textwidth}{!}{%
  \begin{tabular}{m{0.06cm}llcc}
  \toprule
  \multicolumn{2}{c}{\textbf{Task}} & \multicolumn{1}{c}{\textbf{Description}} & \textbf{\makecell[c]{Context \\ Requirement}} & \textbf{Metric} \\
  
  \midrule
  
  \rowcolor{t1}
  \multicolumn{2}{l}{\raisebox{-0.4ex}{\includegraphics[height=1em]{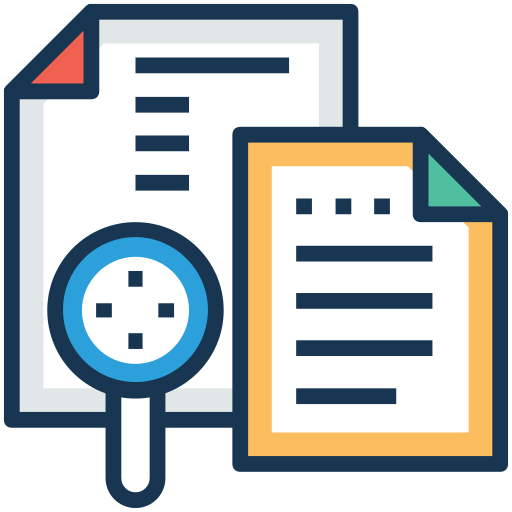}} T1 Retrieval \& Ranking} & \multicolumn{3}{l}{Retrieve content and rank most relevant first} \\
  & T1.1 Global Cohesive Retrieval & Retrieve full text and reorganize & Full & NDCG@k \\
  & T1.2 Key-Snippet Retrieval & Locate target fragment in specified paragraph & Partial & NDCG@k \\
  
  \midrule
  
  \rowcolor{t2}
  \multicolumn{2}{l}{\raisebox{-0.4ex}{\includegraphics[height=1em]{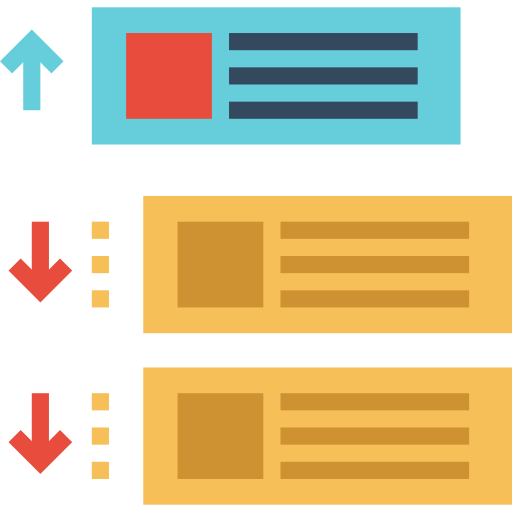}} T2 Sequencing \& Structure Reconstruction} & \multicolumn{3}{l}{Restore timeline or logical order} \\ 
  & T2.1 Global Timeline Reconstruction & Sort unordered events in the whole text & Full & Pairwise Accuracy \\ 
  & T2.2 Local Causal Chain Sorting & Sort content in a specific paragraph & Partial & Pairwise Accuracy \\
  
  \midrule
  
  \rowcolor{t3}
  \multicolumn{2}{l}{\raisebox{-0.4ex}{\includegraphics[height=1em]{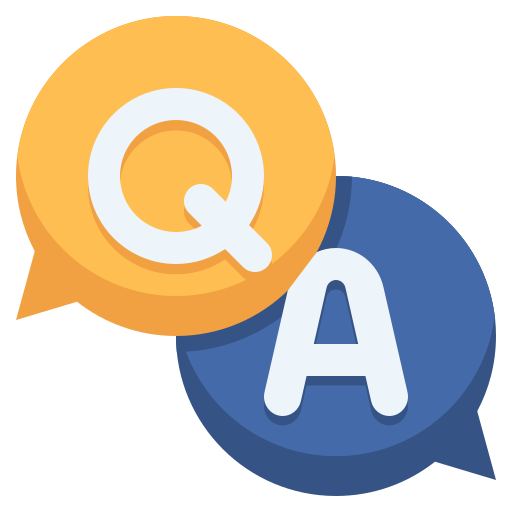}} T3 Evidence-Grounded QA} & \multicolumn{3}{l}{Answer fact/reasoning questions based on evidence} \\
  & T3.1 Multi-Doc Integration QA & Use multi-hop information to answer questions & Full & Accuracy \\
  & T3.2 Single-Hop Fact QA & Answer questions based on local paragraphs & Partial & Accuracy \\
  
  \midrule
  
  \rowcolor{t4}
  \multicolumn{2}{l}{\raisebox{-0.4ex}{\includegraphics[height=1em]{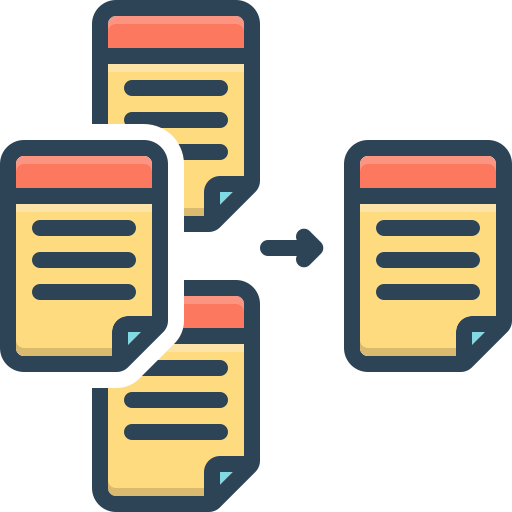}} T4 Summarization \& Synthesis} & \multicolumn{3}{l}{Generate abstract summary under given constraints} \\
  & T4.1 Global-Coverage Constrained Summary & Generate summary of full text & Full & SemSim, ROUGE-L\\
  & T4.2 Query-Focused Summary & Generate summary of specific subtopic & Partial & SemSim, ROUGE-L \\
  
  \midrule
  
  \rowcolor{t5}
  \multicolumn{2}{l}{\raisebox{-0.4ex}{\includegraphics[height=1em]{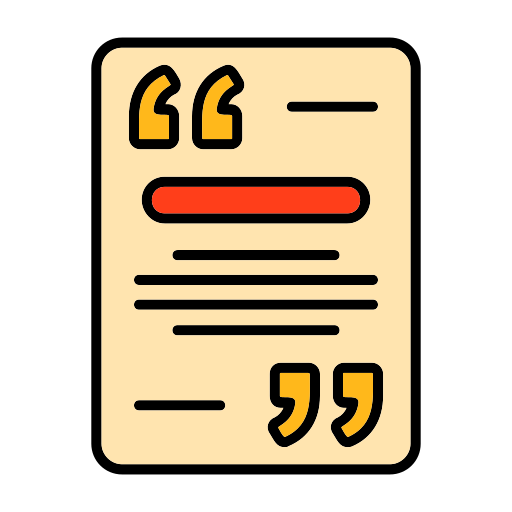}} T5 Attribution \& Citation Alignment} & \multicolumn{3}{l}{Bind correct sources to generated text} \\
  & T5.1 Full-Sentence Citation Alignment & Citation alignment for all sentences & Full & F1 \\
  & T5.2 Key-Statement Citation Alignment & Citation alignment for specified sentences & Partial & F1 \\
  
  \midrule
  
  \rowcolor{t6}
  \multicolumn{2}{l}{\raisebox{-0.4ex}{\includegraphics[height=1em]{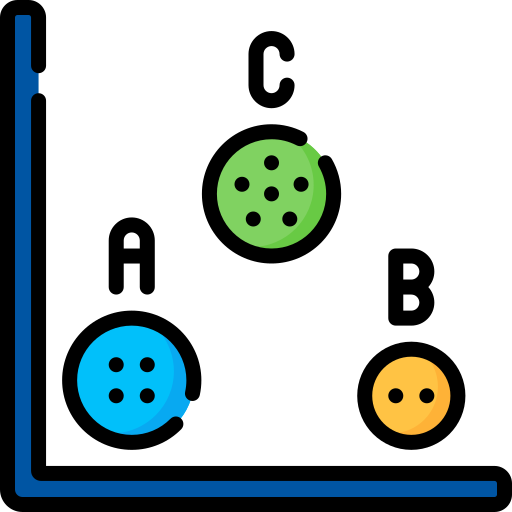}} T6 Aggregation \& Clustering} & \multicolumn{3}{l}{Cluster and output statistics/examples/sort} \\
  & T6.1 Large-Scale Document Clustering & Return all category proportions & Full & SubEM \\
  & T6.2 Targeted Subset Cluster Identification & Return query category instances & Partial & F1 \\
  & T6.3 Global Frequency Analysis & Count and sort global word frequency & Full & Pairwise Accuracy \\
  
  \midrule
  
  \rowcolor{t7}
  \multicolumn{2}{l}{\raisebox{-0.4ex}{\includegraphics[height=1em]{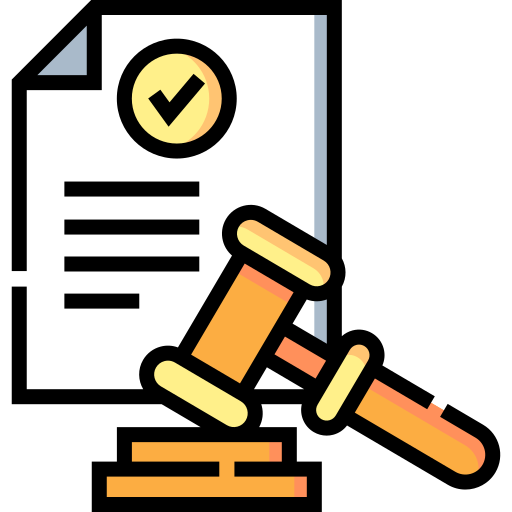}} T7 Consistency \& Compliance Checking} & \multicolumn{3}{l}{Detect and locate contradictions/violations} \\
  & T7.1 Global Conflict \& Inconsistency Localization & Locate contradictory segments in the full text & Full & F1 \\
  & T7.2 Targeted Rule or Condition Violation Detection & Locate content that violates specific rules & Partial & F1 \\
  & T7.3 Comprehensive Error \& Anomaly Sweep & Locate spelling errors in the full text & Full & F1 \\
  
  \midrule
  
  \rowcolor{t8}
  \multicolumn{2}{l}{\raisebox{-0.4ex}{\includegraphics[height=1em]{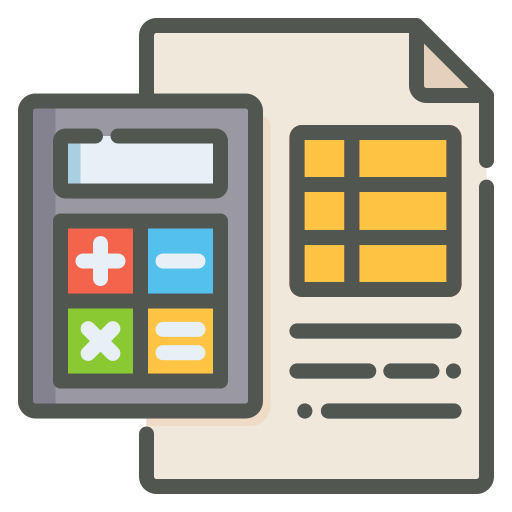}} T8 Structured \& Numeric Reasoning} & \multicolumn{3}{l}{Numerical calculations in structured text} \\
  & T8.1 Structured Multi-Source Consistency Verification & Numerical computation in multi-source & Full & SubEM \\
  & T8.2 Single-Source Targeted Aggregation & Query computation in single-source & Partial & SubEM \\
  & T8.3 Long-Context Procedural State Tracking & Track entity state evolution & Full & SubEM \\
  
  \midrule
  
  \rowcolor{t9}
  \multicolumn{2}{l}{\raisebox{-0.4ex}{\includegraphics[height=1em]{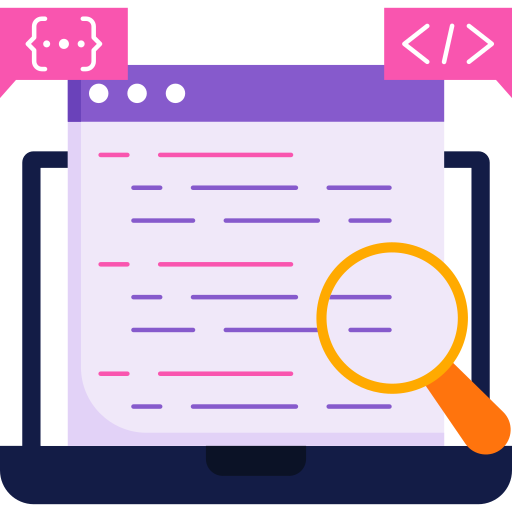}} T9 Version \& Code Diff Analysis} & \multicolumn{3}{l}{Compare changes in different text/code versions} \\
  & T9.1 Dependency-Aware Multi-Version Impact Analysis & Track dependency changes across versions & Full & F1 \\
  & T9.2 Localized Interface Change Detection & Detect local version differences & Partial & F1 \\
  
  \midrule
  
  \rowcolor{t10}
  \multicolumn{2}{l}{\raisebox{-0.4ex}{\includegraphics[height=1em]{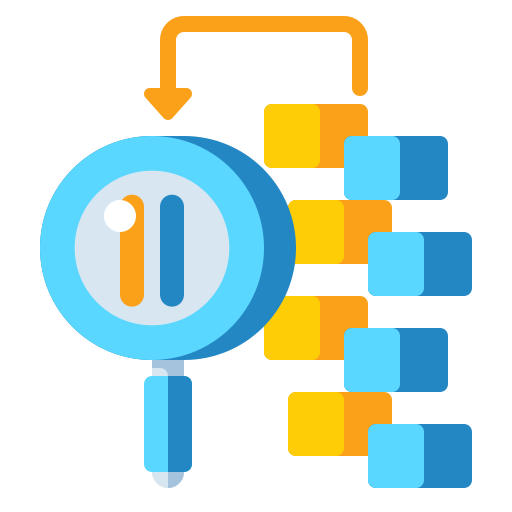}} T10 Rule Induction \& In-Context Learning} & \multicolumn{3}{l}{Summarize rules and make decisions on new samples} \\
  & T10.1 Large-Scale In-Context Rule Induction & Induce rules from the global context & Full & SubEM \\
  & T10.2 Targeted Example-Based Rule Induction & Induce rules from the targeted examples & Partial & SubEM \\
  
  \midrule
  
  \rowcolor{t11}
  \multicolumn{2}{l}{\raisebox{-0.4ex}{\includegraphics[height=1em]{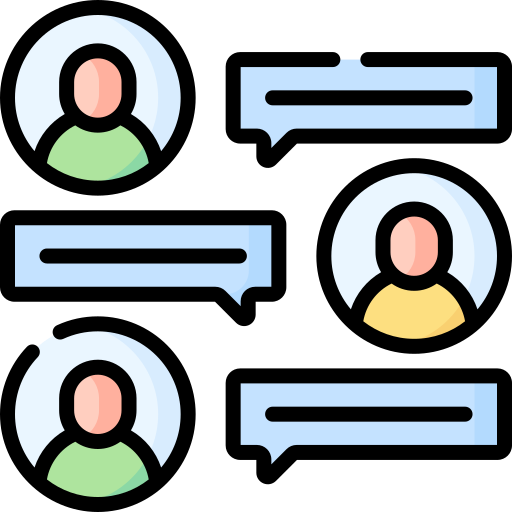}} T11 Dialogue Memory \& Long-Horizon Tracking} & \multicolumn{3}{l}{Track and respond to dialogue history} \\
  & T11.1 Long-Range Entity \& Commitment Tracking & Track entity states across the global context & Full & Accuracy \\
  & T11.2 Short-Range Reference Resolution \& State Query & Resolve references and states in local context & Partial & Accuracy \\
  
  \bottomrule
\end{tabular}}
\caption{Task definitions of LongBench Pro.}
\label{tab:task_definitions}
\end{table*}

\begin{figure*}[h!]
\centering
\includegraphics[width=0.9\textwidth]{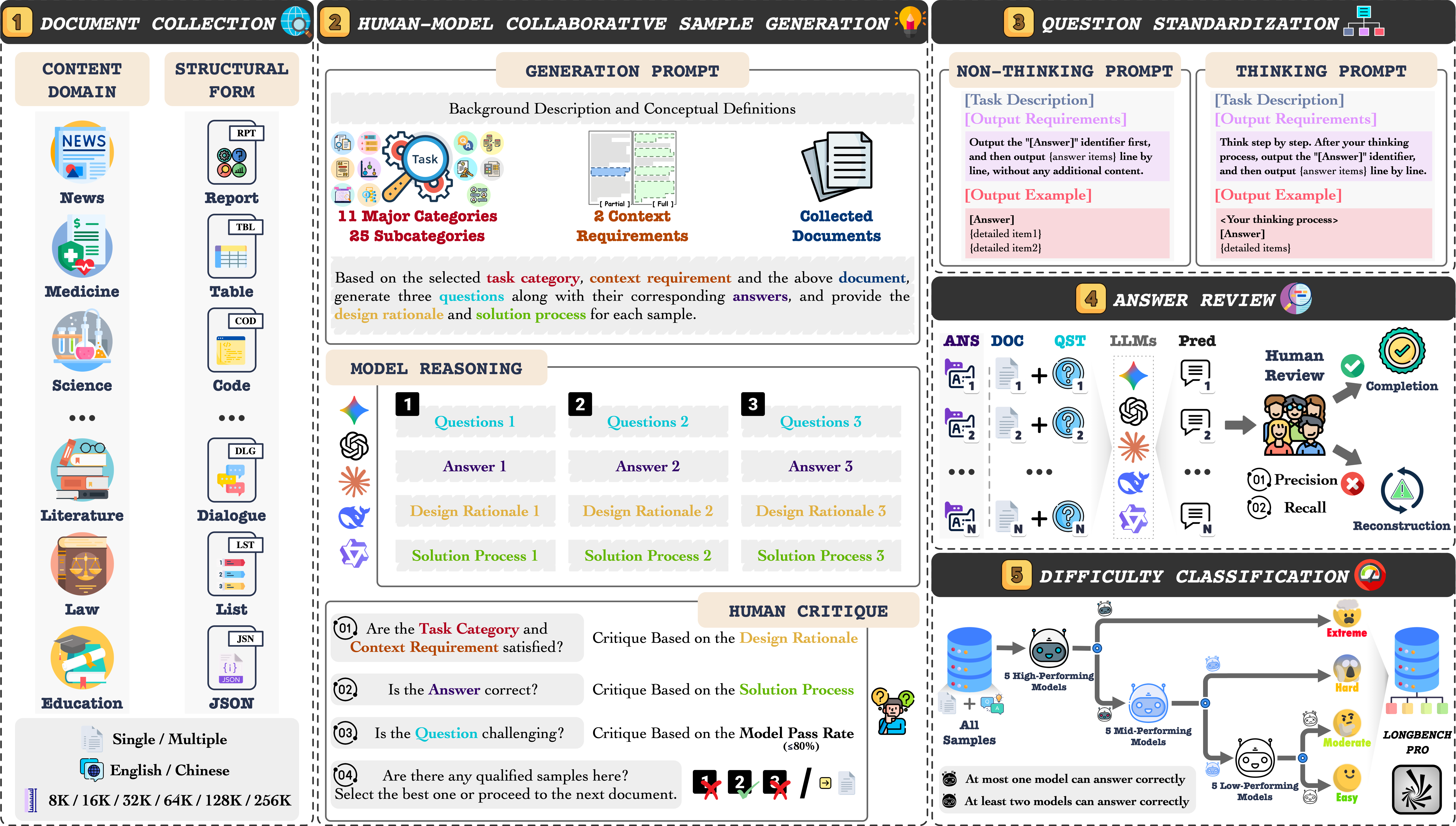}
\caption{The construction process of LongBench Pro includes document collection, human–model collaborative sample generation, question standardization, answer review, and difficulty classification.}
\label{fig:pipeline}
\end{figure*}

\section{Construction Process of LongBench Pro}

\textit{\textbf{Note:}} The prompts and guidelines involved in the construction process are detailed in Appendix~\ref{appendix:annotation_guidelines}.

\subsection{Document Collection}
To ensure realism and coverage, we curate naturally occurring long documents from the public internet across diverse domains (e.g., news, medicine, science, literature, law, and education) and formats (e.g., reports, tables, code, dialogues, lists, and JSON). We balance the collection across single-document and multi-document settings, as well as English and Chinese, and six target length buckets (8k/16k/32k/64k/128k/256k tokens), where token length is measured using the Qwen tokenizer~\cite{qwen3}. Since raw documents rarely match target lengths exactly, we assign a document to a bucket if its length falls within \(\pm 20\%\) of the target. All documents undergo a compliance review by human annotators to exclude content that is privacy-sensitive, copyrighted, or otherwise non-compliant.

\subsection{Human-Model Collaborative Sample Generation}

To balance authenticity with annotation cost, we adopt a human-model collaborative construction strategy. Given a lengthy document, we prompt multiple frontier LLMs (Gemini‑2.5‑Pro~\cite{gemini25}, GPT‑5~\cite{gpt5}, Claude‑4‑Sonnet~\cite{claude4}, DeepSeek‑V3.2~\cite{deepseekv32}, and Qwen3‑235B‑A22B‑Thinking‑2507~\cite{qwen3}) to draft three candidate samples aligned with a target task definition and context requirement, including (i) questions, (ii) reference answers, and (iii) the corresponding design rationales and solution processes to support later verification.

Subsequently, human annotators critically evaluate the model-generated content, focusing primarily on the following aspects:

\begin{enumerate}[label=(\arabic*)]
    \item Verify task alignment and context requirement based on the provided design rationale;
    \item Validate answer correctness using the accompanying solution processes;
    \item Estimate difficulty using the responses of the five drafting models (a sample is considered challenging if at least one model answers incorrectly);
    \item Select the best sample that meets the criteria or can be edited with minimal changes to meet the criteria; if none, move to the next document.
\end{enumerate}

This workflow leverages models for scalable drafting and humans for rigorous verification, mitigating both human cognitive limitations at extreme lengths and model hallucinations. Each accepted sample is reviewed by a long-context expert; failed cases must be revised until they satisfy the criteria. Section~\ref{sec:construction_strategies} empirically evaluates the effectiveness of this construction strategy.

\subsection{Question Standardization}

Chain-of-Thought prompting (CoT)~\cite{cot} demonstrates that the thinking process enhances model performance. To systematically evaluate the upper bound of model capabilities, we construct two rigorously standardized prompt templates for each question: a non-thinking prompt and a thinking prompt. Each prompt includes a task description, output requirements, and an output example, uniformly instructing the model to present answer elements line by line, following the identifier ``[Answer]" for automated extraction and evaluation. The only difference between the two types of prompts is that the non-thinking prompt requires the model to answer directly. In contrast, the thinking prompt requires it to perform explicit step-by-step thinking before producing the final answer.

\subsection{Answer Review}

To ensure sample quality, we systematically review all samples. Annotation experience shows that human annotators are more reliable in judging the correctness of answer components, while models are better at generating diverse candidate components. Based on this complementarity, we first collect predictions from five advanced models for each sample. Then, we instruct annotators to examine each component of the original answer to ensure precision, followed by reviewing model predictions to improve recall. Two annotators independently verify each sample. Samples without detected issues are directly included in the benchmark. If either annotator identifies a potential problem, an additional long-context expert evaluates the sample and decides whether it requires reconstruction.

\begin{figure*}[t!]
\centering
\includegraphics[width=1\textwidth]{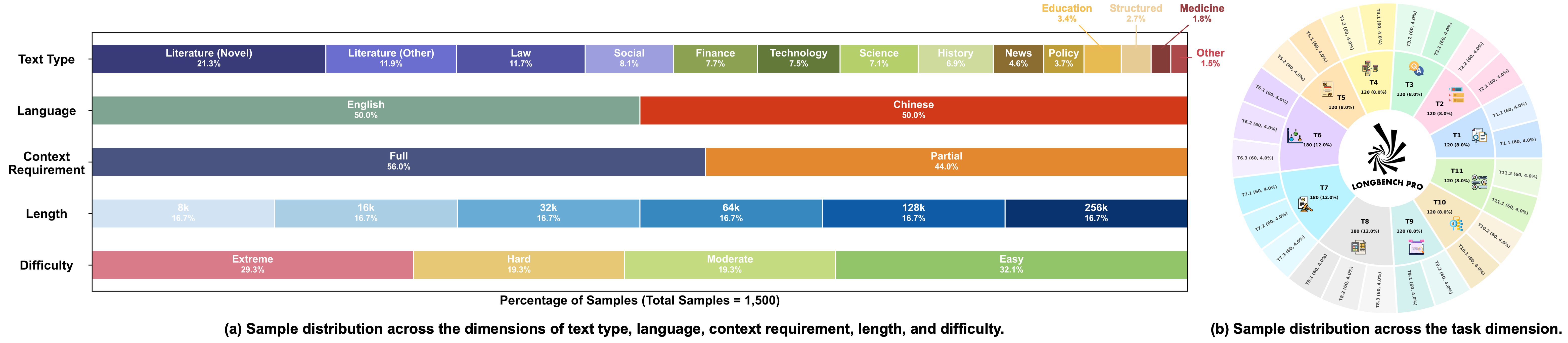}
\caption{Overview of LongBench Pro sample distributions.}
\label{fig:data_source_distribution}
\end{figure*}

\subsection{Difficulty Classification}

To improve the benchmark's utility in real-world applications, we assign each sample a difficulty label defined from a model-centric perspective rather than subjective human ratings~\cite{longbenchv2}, which aligns more closely with the practical needs of contemporary LLM evaluation and provides a more natural foundation for the co-evolution 
of benchmarks and model capabilities. Concretely, we rank models by overall performance and partition them into three tiers (high/mid/low). Within each tier, we select five representative models that perform the best while covering diverse architectures, which reduces sensitivity to outliers and avoids the bias introduced by relying on a single model. This tiered design provides multiple decision boundaries for fine-grained difficulty labeling. The selected models are:

\begin{itemize}
    \item \textbf{High-performing models}: Gemini-2.5-Pro~\cite{gemini25}, GPT-5~\cite{gpt5}, Claude-4-Sonnet~\cite{claude4}, DeepSeek-V3.2~\cite{deepseekv32}, Qwen3-235B-A22B-Thinking-2507~\cite{qwen3}

    \item \textbf{Mid-performing models}: GLM-4.6~\cite{glm46}, DeepSeek-V3-0324~\cite{deepseekv3}, Kimi-K2-Instruct-0905~\cite{kimik2}, Qwen3-30B-A3B-Instruct-2507~\cite{qwen3}, MiniMax-M2~\cite{minimaxm2}

    \item \textbf{Low-performing models}: Ministral-3-8B-Instruct-2512~\cite{mistral3}, Qwen3-8B~\cite{qwen3}, Qwen2.5-72B-Instruct~\cite{qwen25}, Llama-3.1-405B-Instruct~\cite{llama31}, Gemma-3-27B-It~\cite{gemma3}
\end{itemize}

On this basis, we divide the samples into four difficulty levels according to the answering performance of three groups of models, defined as follows:

\begin{itemize}
    \item \textbf{Extreme}: samples that at most one high-performing model can answer correctly (for which a score greater than 0.65 on the summarization task is considered correct);
    
    \item \textbf{Hard}: after excluding Extreme samples, samples that at most one mid-performing model can answer correctly;
    
    \item \textbf{Moderate}: after further excluding Hard samples, samples that at most one low-performing model can answer correctly;
    
    \item \textbf{Easy}: the remaining samples are automatically assigned to this level.
\end{itemize}

This multi-tier progressive approach yields fine-grained difficulty labels aligned with model capabilities and provides a unified, scalable framework for analyzing cross-difficulty performance.

\section{Data Statistics and Validation of LongBench Pro}

We construct LongBench Pro with a balanced design: 5 samples for each combination of 25 secondary tasks, 2 languages, and 6 length buckets, resulting in 1,500 samples in total. Figure~\ref{fig:data_source_distribution} reports the distributions of text type, language, context requirement, length, and difficulty, as well as the per-task composition. To validate sample quality, we uniformly select 300 samples across secondary tasks, languages, and lengths and audit (i) attribute correctness (whether language, length, secondary task, and context requirement are all correct) and (ii) answer correctness (whether the answer is fully correct). Among the 300 samples, the attribute correctness reaches 99.3\%, and the answer correctness reaches 97.3\%, with problematic samples exhibiting only minor deviations (impacting the overall score by only 0.96), demonstrating the high quality of the benchmark samples.

\section{Evaluation}

\subsection{Evaluation Settings}

\textbf{Evaluation Metrics:} We use task-specific metrics summarized in Table~\ref{tab:task_definitions}. T1 (Retrieval \& Ranking) is evaluated by NDCG@k. T2 and T6.3 (ordering-style tasks) use pairwise accuracy based on rank consistency. T3 and T11 are multiple-choice and use accuracy. For tasks with potentially multiple answer components extracted from the source text (T5, T6.2, T7, and T9), we use F1 to penalize spurious components. For tasks with a single canonical answer that is not directly copied from the source (T6.1, T8, and T10), we use SubEM. For T4 (Summary), we combine semantic similarity (SemSim) and ROUGE-L to balance semantic faithfulness and coverage. Each summarization sample includes three reference summaries. The metrics are first computed between the generated summary and each reference summary individually, and the maximum value for each metric is taken to reflect consistency with the best-matching reference. The final weighted score is calculated as:

\begin{small}
\begin{equation}
\begin{aligned}
\text{Score}_{\text{summary}} 
    & = 0.5 \cdot \max_i \text{SemSim}(S_{\text{gen}}, S_{\text{ref}_i}) \\
    & + 0.5 \cdot \max_i \text{ROUGE-L}(S_{\text{gen}}, S_{\text{ref}_i})
\end{aligned}
\end{equation}
\end{small}

All metrics have a value range of \([0,1]\). We report the average score over all samples and multiply by 100.

\noindent\textbf{Evaluated Models:} We evaluate 46 long-context models that vary in transparency (closed-source, such as GPT-5~\cite{gpt5}; open-source, such as GPT-OSS-120B~\cite{gptoss}), thinking mode (thinking, such as DeepSeek-R1~\cite{deepseekr1}; mixed-thinking, such as DeepSeek-V3.2~\cite{deepseekv32}; non-thinking, such as DeepSeek-V3-0324~\cite{deepseekv3}), size (3B, such as Ministral-3-3B-Instruct-2512~\cite{mistral3}; 1T, such as Kimi-K2-Instruct-0905~\cite{kimik2}), architecture (dense, such as Qwen3-32B~\cite{qwen3}; MoE, such as Qwen3-235B-A22B-Instruct-2507~\cite{qwen3}), and context length (128k, such as Gemma-3-27B-It~\cite{gemma3}; 1M, such as Gemini-2.5-Pro~\cite{gemini25}), with the goal of comprehensively assessing the long-context performance of current LLMs.

\noindent\textbf{Inference Settings:} We uniformly use each model's default inference parameters to run inference three times, and report both the general performance (the average score across multiple responses) and the upper-bound performance (Best-of-N, the highest score among the multiple responses; Pass@N, the probability that at least one response is completely correct). For models without default inference parameters, we set the temperature to 1.0. For thinking models, we use the non-thinking prompt and report their thinking scores. For mixed-thinking models, we use non-thinking prompts and report both the non-thinking and thinking scores under the disabled and enabled thinking states. For non-thinking (instruct) models, we use non-thinking prompts and thinking prompts separately to report the corresponding non-thinking and thinking scores. For the thinking score, models that support a 256k context length set the output length to 32k to enable more thorough reasoning, while for other models, we set the output length to 8k to reserve more budget for the input. For the non-thinking score, the output length is uniformly set to 1k. When the sample length exceeds the model's context length, we truncate the sample from the middle to an appropriate length for input, with the truncation length uniformly set to the model's context length minus the output length. Detailed inference parameter settings for different models are provided in Appendix~\ref{appendix:inference_parameter_settings}. Unless otherwise specified, we report the thinking scores by default.

\subsection{General Performance}

Table~\ref{tab:general_performance} summarizes the general performance of 46 models on LongBench Pro (gray-shaded cells are thinking scores). We observe a clear stratification in overall long-context performance. The top three models are Gemini-2.5-Pro (73.42), GPT-5 (72.61), and Claude-4-Sonnet (69.87). Among open-source models, DeepSeek-V3.2 (67.82) and Qwen3-235B-A22B-Thinking-2507 (66.97) are the strongest, narrowing the gap to the best closed-source model to within 6 points.

\begin{table*}[t]
\centering
\resizebox{\textwidth}{!}{%
\begin{tabular}{cl | c c | c >{\columncolor{gray!20}}c | c>{\columncolor{gray!20}}c c>{\columncolor{gray!20}}c | c>{\columncolor{gray!20}}c c>{\columncolor{gray!20}}c c>{\columncolor{gray!20}}c c>{\columncolor{gray!20}}c}
\toprule
\multicolumn{2}{c|}{\multirow{2}{*}{\textbf{Model}}} &
\textbf{Model} &
\textbf{Context} &
\multicolumn{2}{c|}{\multirow{2}{*}{\textbf{Overall}}} &
\multicolumn{4}{c|}{\textbf{Language}} &
\multicolumn{8}{c}{\textbf{Difficulty}} \\

\multicolumn{2}{c|}{} & 
\textbf{Type} & 
\textbf{Length} &
\multicolumn{2}{c|}{} &
\multicolumn{2}{c}{\textbf{English}} &
\multicolumn{2}{c|}{\textbf{Chinese}} &
\multicolumn{2}{c}{\textbf{Extreme}} &
\multicolumn{2}{c}{\textbf{Hard}} &
\multicolumn{2}{c}{\textbf{Moderate}} &
\multicolumn{2}{c}{\textbf{Easy}} \\

\midrule

\multirow{5}{*}{\includegraphics[height=1.5em]{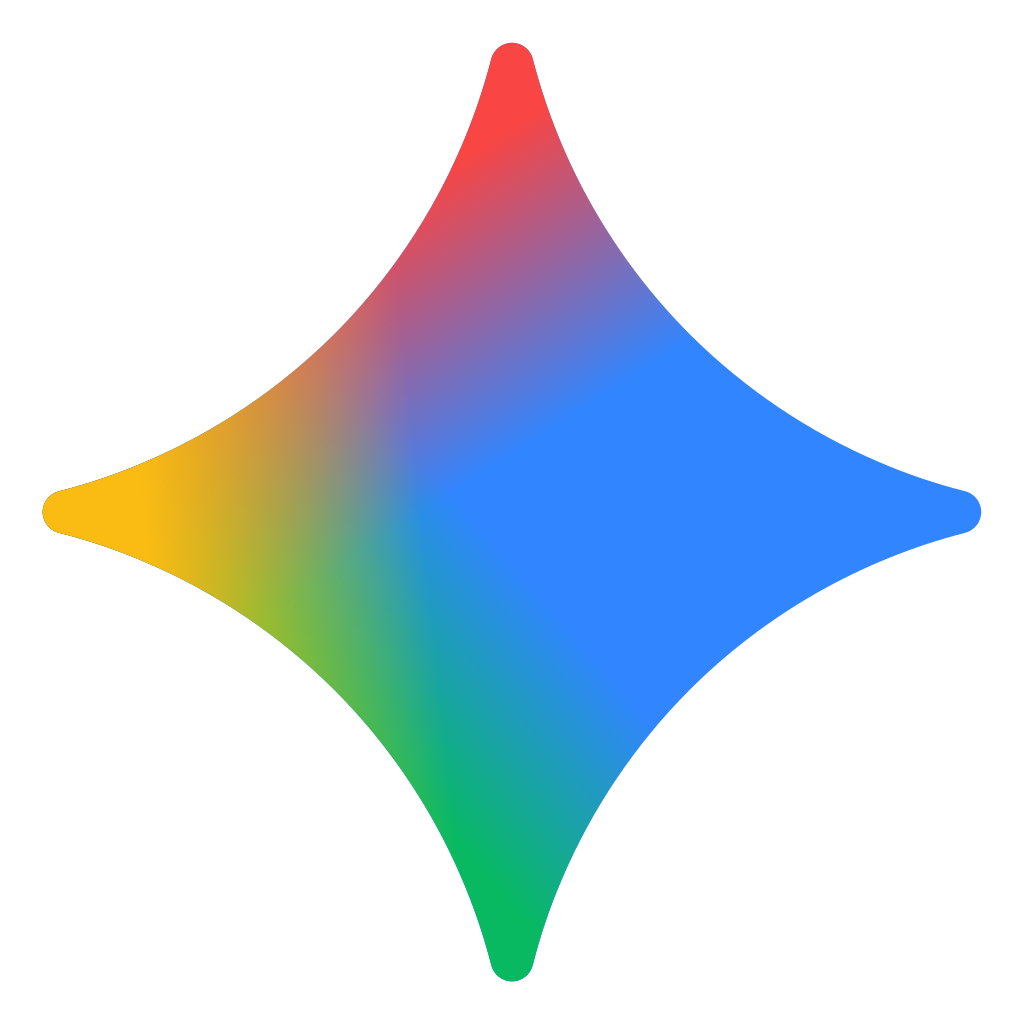}}

& Gemini-2.5-Pro & Thinking & 1M 
& - & \textcolor{red!70}{\textbf{73.42}} & - & \textcolor{green!70!black}{\textbf{72.35}} & - & \textcolor{red!70}{\textbf{74.49}} & - & \textcolor{red!70}{\textbf{50.77}} & - & \textcolor{red!70}{\textbf{81.03}} & - & \textcolor{green!70!black}{\textbf{81.98}} & - & 84.40 \\

& Gemini-2.5-Flash & Mixed & 1M 
& \textcolor{green!70!black}{\textbf{55.92}} & 67.41 & \textcolor{green!70!black}{\textbf{55.29}} & 67.22 & \textcolor{red!70}{\textbf{56.54}} & 67.59 & \textcolor{red!70}{\textbf{44.26}} & \textcolor{blue!70}{\textbf{47.39}} & \textcolor{red!70}{\textbf{57.87}} & 72.19 & \textcolor{red!70}{\textbf{53.99}} & 72.39 & \textcolor{blue!70}{\textbf{66.55}} & 79.82 \\

& Gemma-3-27B-It & Instruct & 128k 
& 36.14 & 37.34 & 37.46 & 40.89 & 34.81 & 33.78 & 30.22 & 27.78 & 33.20 & 30.56 & 25.04 & 24.53 & 49.96 & 57.81 \\

& Gemma-3-12B-It & Instruct & 128k 
& 32.16 & 31.92 & 33.03 & 34.43 & 31.28 & 29.41 & 26.44 & 25.74 & 30.43 & 28.02 & 23.39 & 22.61 & 43.66 & 45.48 \\

& Gemma-3-4B-It & Instruct & 128k 
& 21.76 & 21.20 & 22.63 & 23.28 & 20.89 & 19.12 & 19.31 & 18.72 & 20.70 & 19.87 & 15.82 & 13.85 & 28.18 & 28.66 \\

\midrule

\multirow{4}{*}{\includegraphics[height=1.5em]{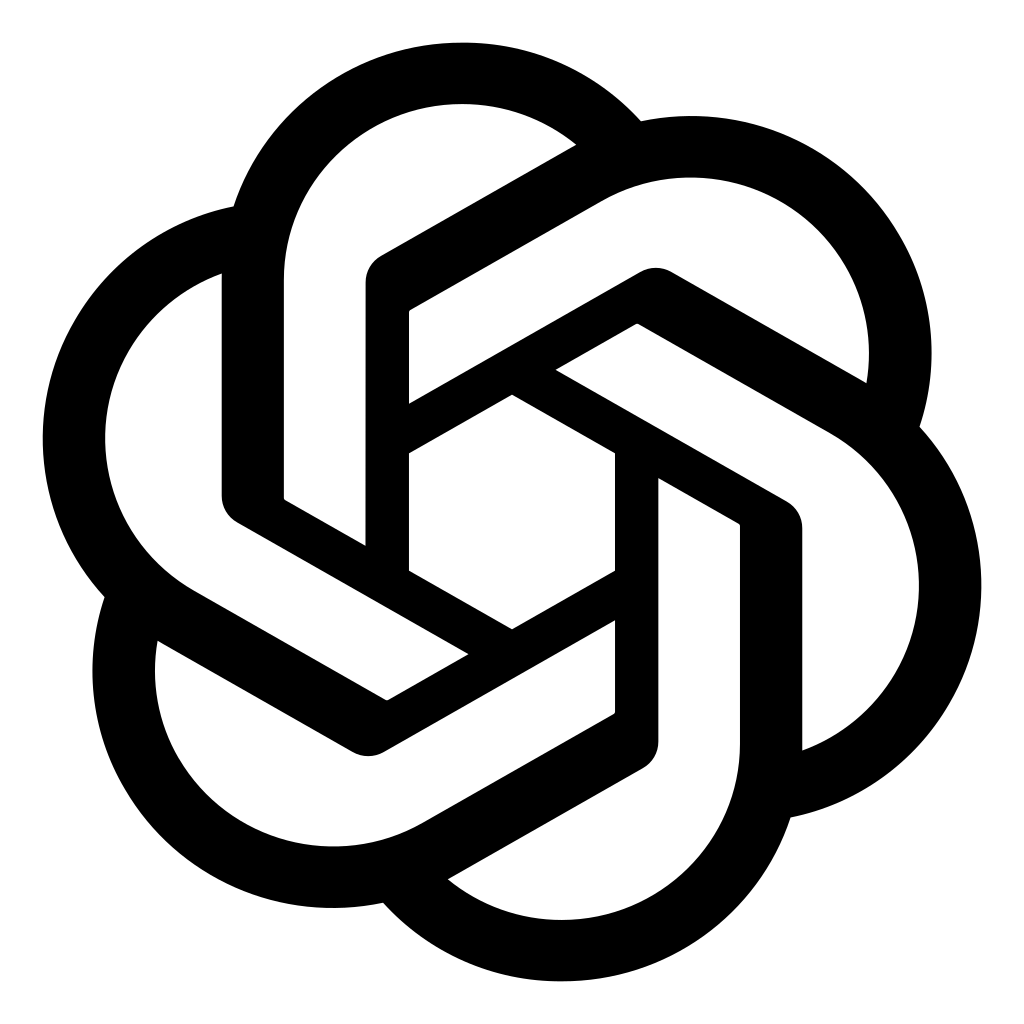}} 

& GPT-5 & Thinking & 272k & - & \textcolor{green!70!black}{\textbf{72.61}} & - & \textcolor{red!70}{\textbf{73.24}} & - & \textcolor{green!70!black}{\textbf{71.97}} & - & \textcolor{green!70!black}{\textbf{48.37}} & - & \textcolor{green!70!black}{\textbf{78.74}} & - & \textcolor{red!70}{\textbf{82.31}} & - & \textcolor{green!70!black}{\textbf{85.23}} \\

& GPT-4o & Instruct & 128k & 46.67 & 49.44 & 47.67 & 52.61 & 45.66 & 46.26 & 36.30 & 34.39 & 44.88 & 41.35 & 43.03 & 43.07 & 59.38 & 71.84 \\

& GPT-OSS-120B & Thinking & 128k & - & 52.61 & - & 54.67 & - & 50.54 & - & 35.4 & - & 44.97 & - & 50.66 & - & 74.06 \\

& GPT-OSS-20B & Thinking & 128k & - & 44.66 & - & 47.83 & - & 41.49 & - & 31.59 & - & 35.89 & - & 39.33 & - & 65.05 \\

\midrule

\multirow{2}{*}{\includegraphics[height=1.5em]{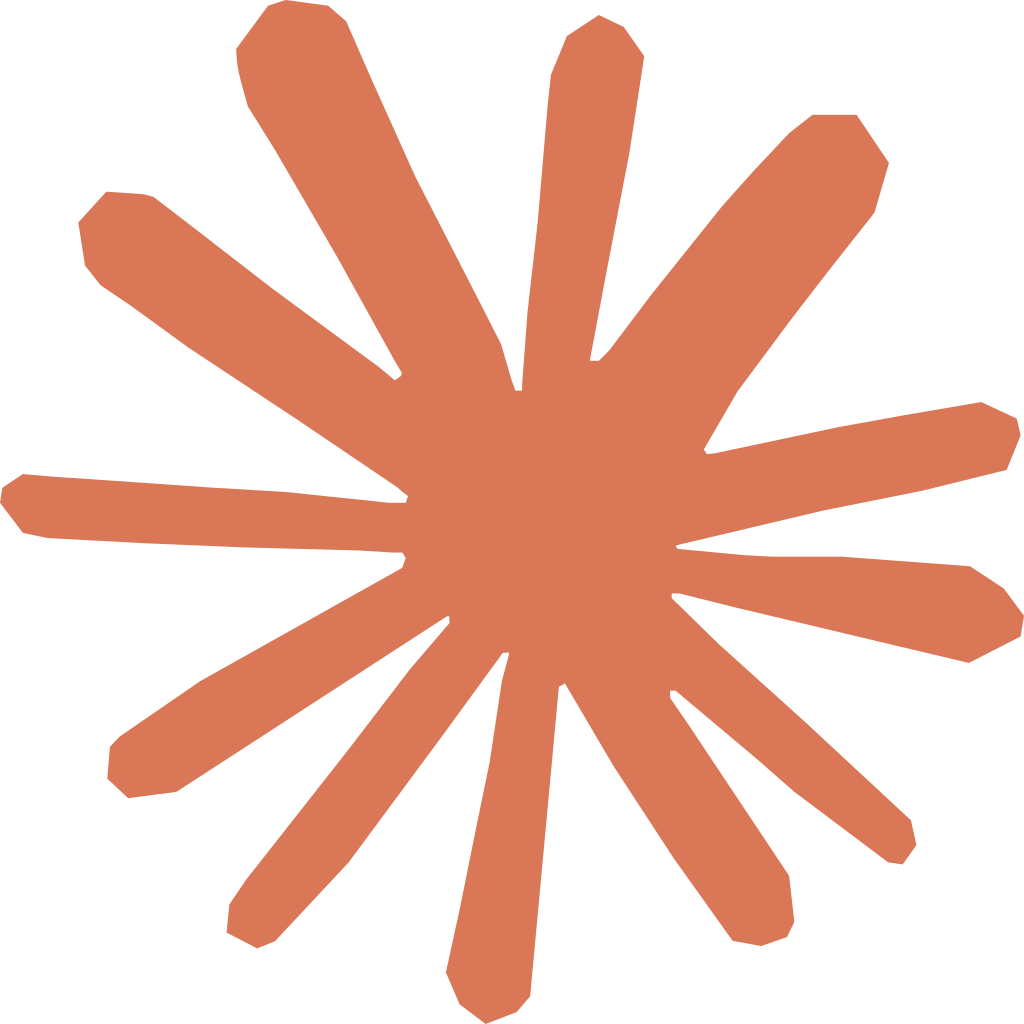}} 

& Claude-4-Sonnet & Mixed & 1M 
& \textcolor{red!70}{\textbf{56.07}} & \textcolor{blue!70}{\textbf{69.87}} & \textcolor{red!70}{\textbf{57.14}} & \textcolor{blue!70}{\textbf{71.09}} & \textcolor{green!70!black}{\textbf{54.99}} & \textcolor{blue!70}{\textbf{68.65}} & \textcolor{green!70!black}{\textbf{42.92}} & 47.05 & \textcolor{green!70!black}{\textbf{57.57}} & \textcolor{blue!70}{\textbf{74.72}} & \textcolor{green!70!black}{\textbf{53.96}} & \textcolor{blue!70}{\textbf{76.58}} & \textcolor{green!70!black}{\textbf{68.42}} & 83.78 \\

& Claude-3.7-Sonnet & Mixed & 200k 
& 51.45 & 59.66 & 51.89 & 60.49 & 51.00 & 58.84 & 37.31 & 40.07 & 47.29 & 56.58 & 48.38 & 61.56 & \textcolor{red!70}{\textbf{68.69}} & 78.26 \\

\midrule

\multirow{5}{*}{\includegraphics[height=1.5em]{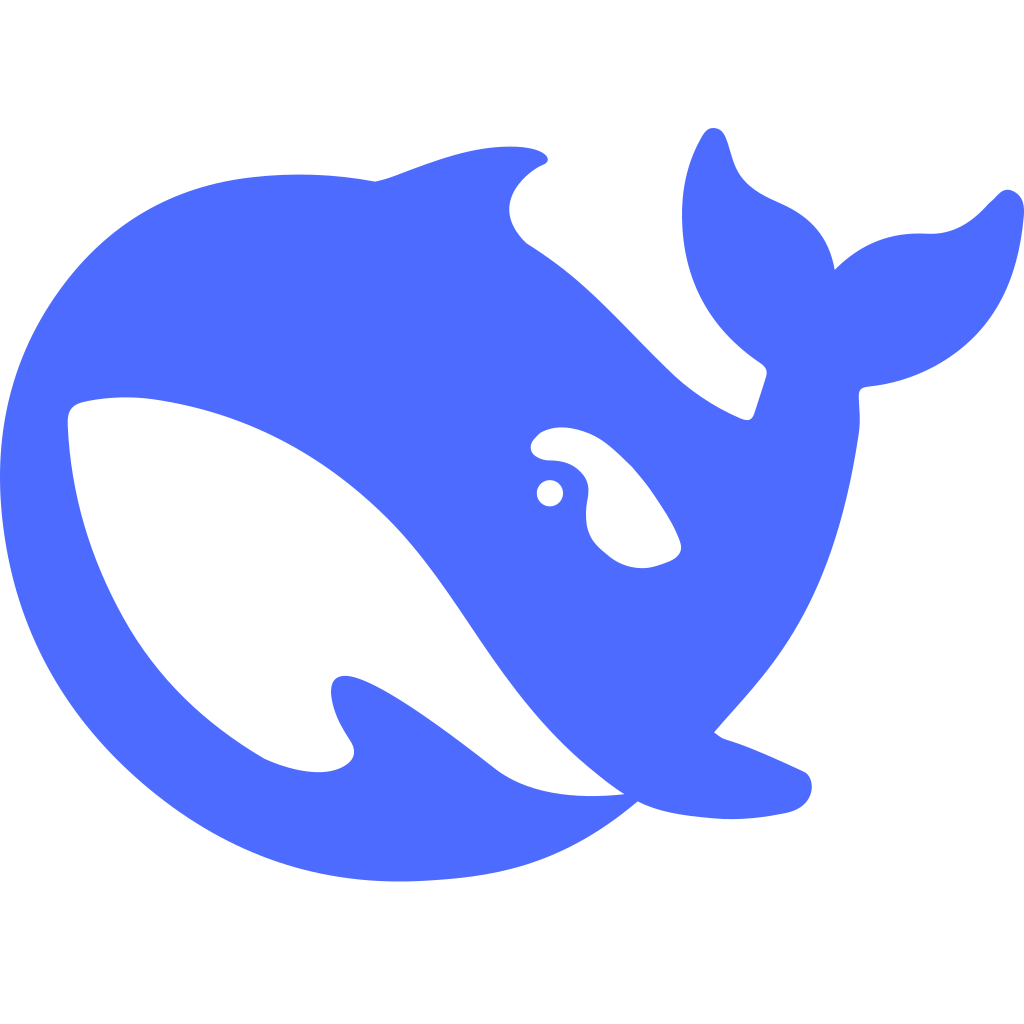}} 

& DeepSeek-V3.2 & Mixed & 160k 
& 51.67 & 67.82 & 50.61 & 67.89 & 52.73 & 67.75 & 40.45 & 44.27 & 51.63 & 67.73 & 51.36 & 75.08 & 62.12 & \textcolor{blue!70}{\textbf{85.02}} \\

& DeepSeek-V3.1 & Mixed & 128k 
& 51.39 & 66.22 & 50.35 & 66.17 & 52.42 & 66.26 & 41.07 & 42.68 & 49.29 & 62.22 & 48.80 & 73.53 & 63.61 & \textcolor{red!70}{\textbf{85.72}} \\

& DeepSeek-R1-0528 & Thinking & 128k 
& - & 61.89 & - & 59.90 & - & 63.89 & - & 41.49 & - & 53.68 & - & 66.53 & - & 82.67 \\

& DeepSeek-R1 & Thinking & 128k 
& - & 60.07 & - & 62.00 & - & 58.13 & - & 40.76 & - & 53.39 & - & 58.83 & - & 82.44 \\

& DeepSeek-V3-0324 & Instruct & 128k 
& 51.70 & 56.71 & 51.62 & 58.14 & 51.78 & 55.27 & 40.40 & 38.69 & 48.30 & 46.20 & 49.68 & 57.14 & 65.26 & 79.20 \\

\midrule

\multirow{13}{*}{\includegraphics[height=1.5em]{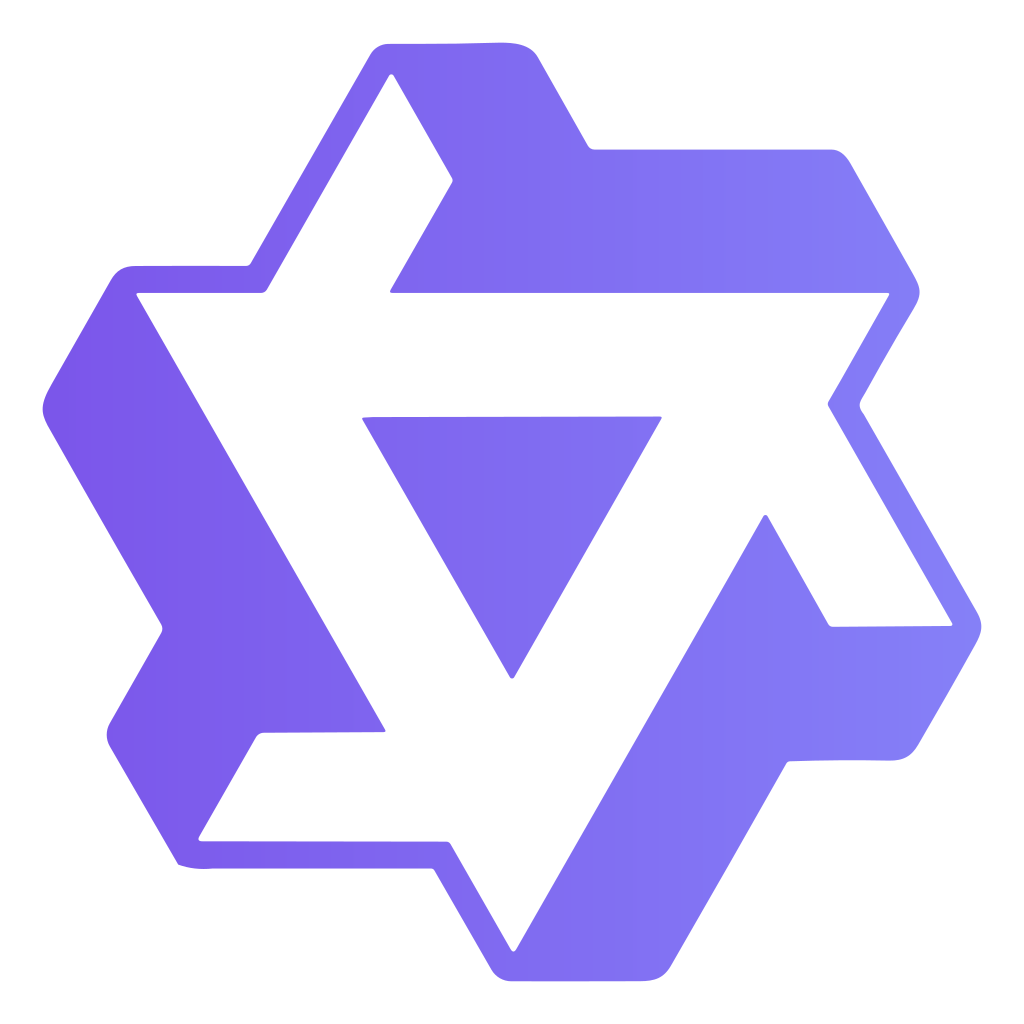}} 
& Qwen3-235B-A22B-Thinking-2507 & Thinking & 256k 
& - & 66.97 & - & 66.83 & - & 67.12 & - & 43.39 & - & 67.10 & - & 75.12 & - & 83.55 \\

& Qwen3-235B-A22B-Instruct-2507 & Instruct & 256k 
& \textcolor{blue!70}{\textbf{52.51}} & 63.77 & \textcolor{blue!70}{\textbf{52.22}} & 63.88 & \textcolor{blue!70}{\textbf{52.80}} & 63.65 & \textcolor{blue!70}{\textbf{42.07}} & 43.24 & \textcolor{blue!70}{\textbf{52.34}} & 58.60 & \textcolor{blue!70}{\textbf{53.14}} & 68.15 & 61.76 & 82.98 \\

& Qwen3-Next-80B-A3B-Thinking & Thinking & 256k 
& - & 63.95 & - & 62.91 & - & 64.99 & - & 42.47 & - & 61.46 & - & 69.23 & - & 81.90 \\

& Qwen3-Next-80B-A3B-Instruct & Instruct & 256k 
& 51.54 & 60.76 & 50.39 & 60.30 & 52.69 & 61.22 & 39.39 & 40.47 & 49.93 & 54.74 & 48.77 & 64.16 & 65.25 & 80.84 \\

& Qwen3-30B-A3B-Thinking-2507 & Thinking & 256k 
& - & 59.68 & - & 60.14 & - & 59.22 & - & 40.47 & - & 52.76 & - & 62.55 & - & 79.64 \\

& Qwen3-30B-A3B-Instruct-2507 & Instruct & 256k 
& 43.84 & 54.52 & 43.17 & 55.55 & 44.5 & 53.49 & 35.44 & 37.05 & 41.32 & 44.04 & 39.04 & 56.47 & 55.89 & 75.59 \\

& Qwen3-4B-Thinking-2507 & Thinking & 256k 
& - & 50.10 & - & 49.81 & - & 50.39 & - & 35.31 & - & 40.99 & - & 47.66 & - & 70.53 \\

& Qwen3-4B-Instruct-2507 & Instruct & 256k 
& 36.78 & 45.68 & 35.70 & 46.27 & 37.85 & 45.10 & 30.29 & 31.09 & 33.94 & 36.96 & 30.21 & 39.69 & 48.33 & 67.82 \\

& Qwen3-32B & Mixed & 128k 
& 40.28 & 51.12 & 39.78 & 52.24 & 40.77 & 50.01 & 32.56 & 36.45 & 38.61 & 42.24 & 34.30 & 46.18 & 51.90 & 72.80 \\

& Qwen3-14B & Mixed & 128k 
& 37.11 & 47.14 & 36.61 & 50.53 & 37.61 & 43.75 & 31.13 & 33.66 & 35.34 & 38.41 & 29.07 & 39.03 & 48.44 & 69.55 \\

& Qwen3-8B & Mixed & 128k 
& 33.41 & 44.34 & 33.04 & 44.60 & 33.79 & 44.08 & 29.99 & 33.50 & 31.09 & 37.10 & 25.20 & 30.16 & 42.86 & 67.08 \\

& Qwen3-4B & Mixed & 128k 
& 31.26 & 40.82 & 31.60 & 41.94 & 30.92 & 39.70 & 27.20 & 30.69 & 30.10 & 34.07 & 23.33 & 31.27 & 40.43 & 59.85 \\

& Qwen2.5-72B-Instruct & Instruct & 128k 
& 39.64 & 44.09 & 39.48 & 44.99 & 39.79 & 43.18 & 32.36 & 31.71 & 35.90 & 36.45 & 31.36 & 31.03 & 53.48 & 67.80 \\

\midrule

\multirow{2}{*}{\raisebox{-0.5em}{\includegraphics[height=1.5em]{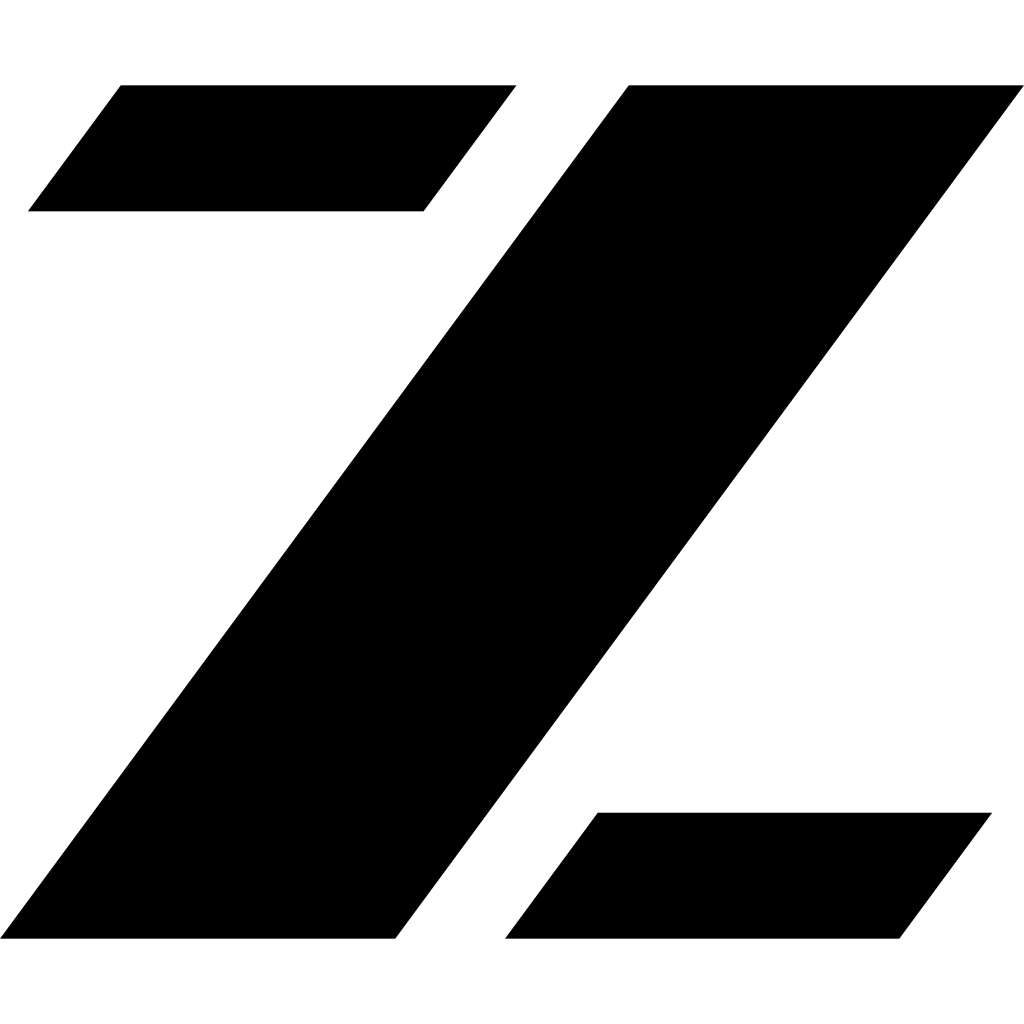}} } 

& GLM-4.6 & Mixed & 198k 
& 45.85 & 58.21 & 45.64 & 56.50 & 46.07 & 59.92 & 37.30 & 38.88 & 43.07 & 48.92 & 40.05 & 60.95 & 58.82 & 79.78 \\

& GLM-4.5 & Mixed & 128k 
& 43.04 & 55.48 & 43.05 & 53.57 & 43.02 & 57.39 & 35.06 & 37.94 & 40.21 & 47.38 & 36.92 & 55.13 & 55.68 & 76.55 \\

\midrule

\raisebox{-0.5em}{\includegraphics[height=1.5em]{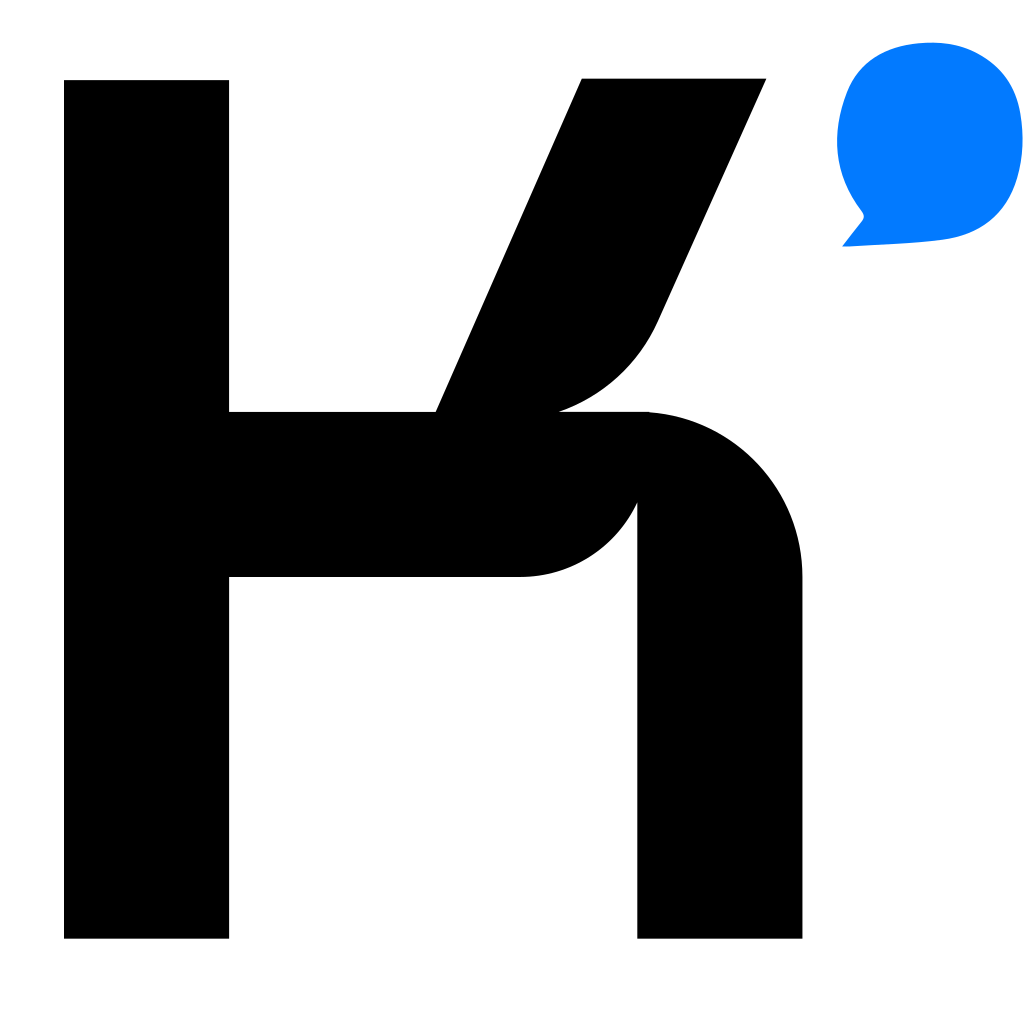}}
& Kimi-K2-Instruct-0905 & Instruct & 256k &  50.09 & 55.53 & 49.90 & 56.96 & 50.29 & 54.10 & 39.61 & 38.25 & 43.43 & 43.75 & 49.05 & 57.33 & 64.92 & 77.29 \\

\midrule

\multirow{2}{*}{\includegraphics[height=1.5em]{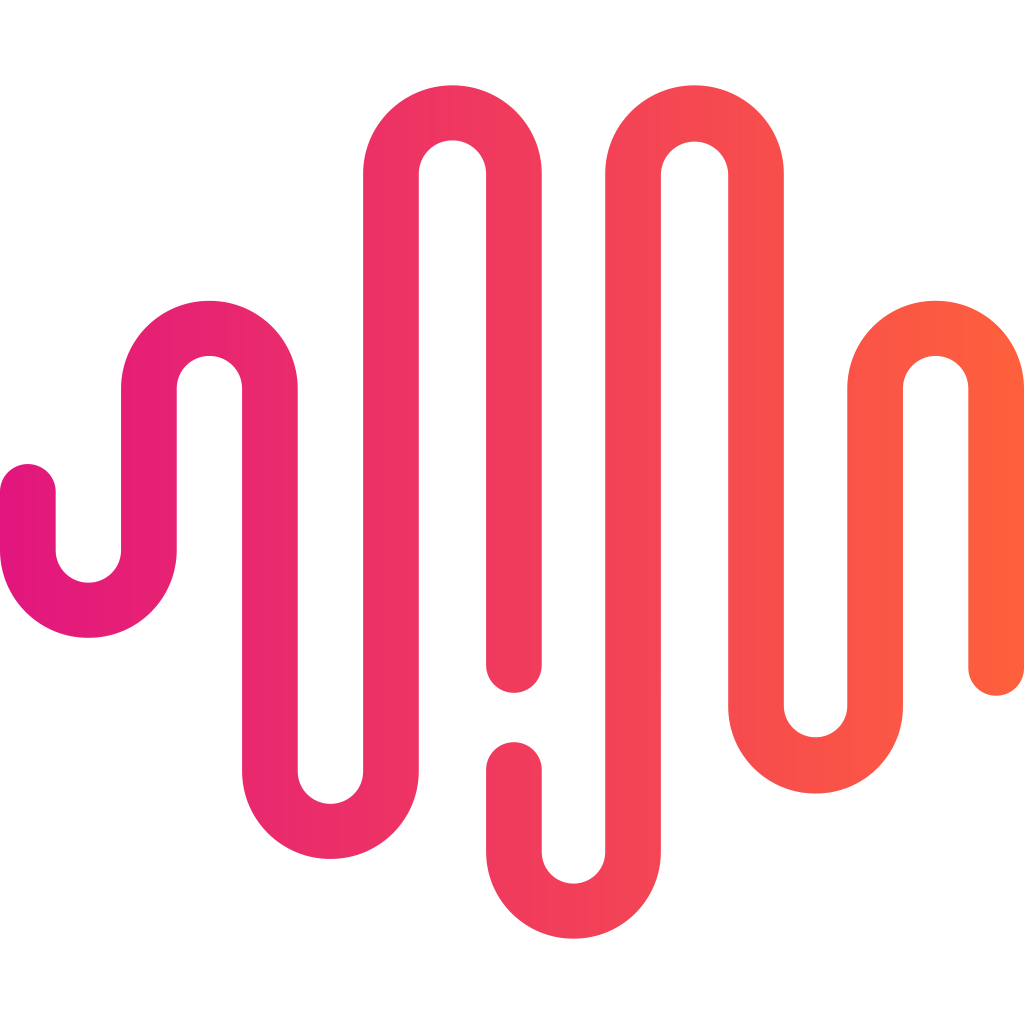}} 
& MiniMax-M2 & Thinking & 192k 
& - & 53.21 & - & 52.87 & - & 53.55 & - & 34.98 & - & 42.58 & - & 59.92 & - & 72.20 \\

& MiniMax-Text-01 & Instruct & 4M 
& 41.14 & 45.00 & 40.21 & 44.17 & 42.06 & 45.82 & 33.57 & 33.78 & 38.67 & 38.02 & 38.23 & 40.82 & 51.26 & 61.92 \\

\midrule

\multirow{7}{*}{\includegraphics[height=1.5em]{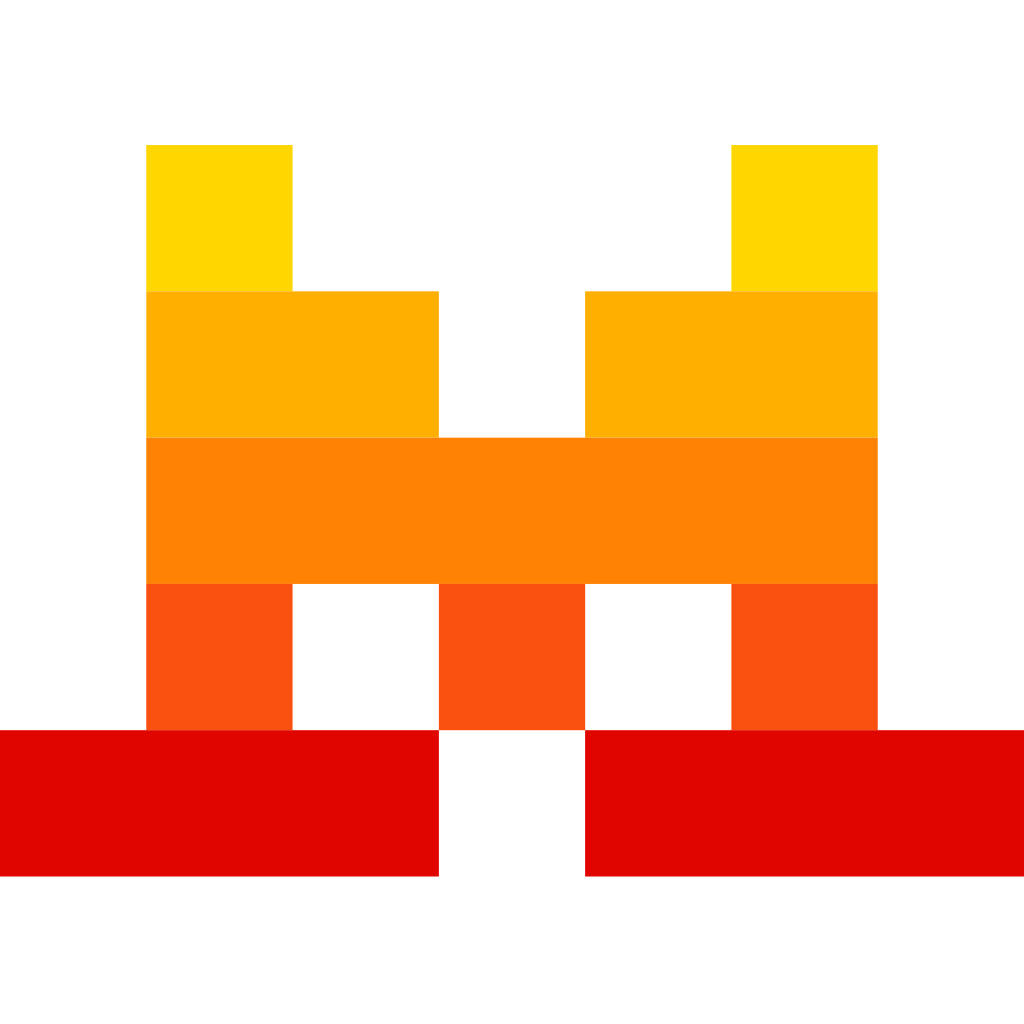}} 

& Ministral-3-14B-Instruct-2512 & Instruct & 256k 
& 40.14 & 45.80 & 39.71 & 47.75 & 40.56 & 43.85 & 33.85 & 31.66 & 35.04 & 37.48 & 34.02 & 39.35 & 52.60 & 67.56 \\

& Ministral-3-8B-Instruct-2512 & Instruct & 256k 
& 37.80 & 44.46 & 36.61 & 46.17 & 39.00 & 42.75 & 31.88 & 31.86 & 32.11 & 34.99 & 31.73 & 35.26 & 50.27 & 67.14 \\

& Ministral-3-3B-Instruct-2512 & Instruct & 256k 
& 30.18 & 34.54 & 27.75 & 36.42 & 32.61 & 32.66 & 25.97 & 26.70 & 28.57 & 30.23 & 23.80 & 25.60 & 38.81 & 49.65 \\

& Magistral-Small-2509 & Thinking & 128k 
& - & 38.40 & - & 40.40 & - & 36.40 & - & 30.52 & - & 32.92 & - & 29.44 & - & 54.25 \\

& Mistral-Small-3.2-24B-Instruct-2506 & Instruct & 128k 
& 37.32 & 39.87 & 39.23 & 42.56 & 35.41 & 37.18 & 31.79 & 29.77 & 33.73 & 31.76 & 27.42 & 27.74 & 50.45 & 61.22 \\

& Mistral-Large-Instruct-2411 & Instruct & 128k 
& 31.69 & 36.25 & 33.10 & 39.14 & 30.28 & 33.36 & 27.39 & 28.65 & 29.88 & 29.42 & 23.42 & 25.62 & 41.66 & 53.65 \\

& Ministral-8B-Instruct-2410 & Instruct & 128k 
& 17.56 & 14.43 & 18.65 & 16.53 & 16.47 & 12.33 & 17.83 & 15.06 & 18.61 & 13.98 & 12.26 & 9.89 & 19.86 & 16.84 \\

\midrule

\multirow{5}{*}{\includegraphics[height=1.5em]{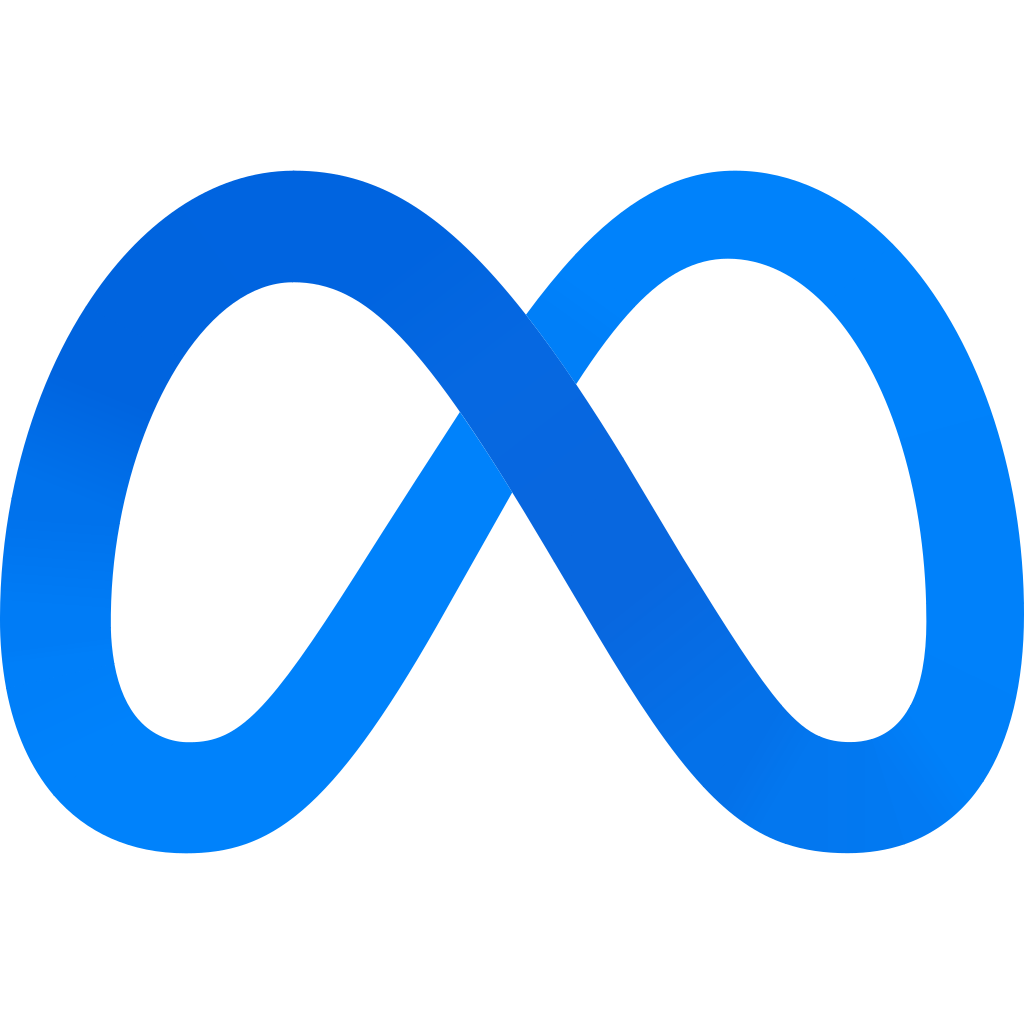}} 

& Llama-3.1-405B-Instruct & Instruct & 128k 
& 40.07 & 40.66 & 42.03 & 44.46 & 38.11 & 36.86 & 33.44 & 29.81 & 35.51 & 34.09 & 29.07 & 29.22 & 55.45 & 61.36 \\

& Llama-3.3-70B-Instruct & Instruct & 128k 
& 31.89 & 33.69 & 35.12 & 39.15 & 28.66 & 28.23 & 26.53 & 24.32 & 29.07 & 28.59 & 22.61 & 22.61 & 44.04 & 51.94 \\

& Llama-3.1-70B-Instruct & Instruct & 128k 
& 31.53 & 32.12 & 35.10 & 36.85 & 27.96 & 27.40 & 26.22 & 23.93 & 28.86 & 28.04 & 21.46 & 21.44 & 44.02 & 48.46 \\

& Llama-3.1-8B-Instruct & Instruct & 128k 
& 21.09 & 20.06 & 24.28 & 25.40 & 17.91 & 14.71 & 21.00 & 19.68 & 21.22 & 17.99 & 13.82 & 12.32 & 25.47 & 26.28 \\

& Llama-3.2-3B-Instruct & Instruct & 128k 
& 15.71 & 12.58 & 20.90 & 16.45 & 10.51 & 8.71 & 16.63 & 15.57 & 15.01 & 10.48 & 10.37 & 7.17 & 18.49 & 14.35 \\

\bottomrule
\end{tabular}}
\caption{General performance on LongBench Pro. \setlength{\fboxsep}{1pt}\colorbox{gray!20}{Gray-shaded} cells represent thinking scores. The best three performance results are highlighted using \textcolor{red!70}{\textbf{red}} (1\textsuperscript{st}), \textcolor{green!70!black}{\textbf{green}} (2\textsuperscript{nd}), and \textcolor{blue!70}{\textbf{blue}} (3\textsuperscript{rd}) font colors, respectively.}
\label{tab:general_performance}
\end{table*}

Based on an in-depth analysis of the evaluation results in Table~\ref{tab:general_performance}, we draw the following insights:

\textbf{(1) Long-Context Optimization Outperforms Model Size Scaling.} For long-context tasks, the Model Size Scaling Law still holds. For example, Qwen3 (32k natively and 128k with YaRN) improves in performance from 4B to 32B (40.82 → 51.12), but its marginal gains decrease. In contrast, the long-context optimized Qwen3-4B-Instruct-2507 (256k) achieves a score of 45.68, surpassing Qwen3-8B (44.34), while Qwen3-30B-A3B-Instruct-2507 (256k) attains a high score of 54.52, outperforming the larger Qwen3-32B (51.12). This demonstrates that extending the effective context length constitutes the primary approach for improving long-context performance, and its effectiveness far exceeds that of scaling up parameters by several times.

\textbf{(2) Discrepancy between Claimed Context Length and Effective Context Length.} For some models, the claimed context length does not positively correlate with their actual performance. For example, although MiniMax-Text-01 claims to support a context length of up to 4M, its overall score is only 45.00, even falling behind most models that have a context length of merely 128k. This clear inconsistency indicates that the ability of a model to accept longer text inputs does not equate to its ability to effectively leverage this information for association, integration, and reasoning. In other words, the claimed context length reflects the model's input capacity, whereas the effective context length reflects its actual long-context understanding and processing capability, and a significant gap may exist between the two.

\textbf{(3) Uneven Distribution of Long-Context Capabilities Between Chinese and English.} Different series of models exhibit clearly uneven performance in long-context tasks across languages. For instance, series such as GPT, Claude, Mistral, and Llama generally perform better in English long-context scenarios, whereas series like GLM, Kimi, and MiniMax demonstrate stronger capabilities in Chinese long-context tasks. This phenomenon indicates that the ``language alignment'' of current LLMs in long-context processing remains insufficient, and language differences significantly affect model robustness and generalization. However, it is noteworthy that as model scale increases and overall capabilities improve, the performance gap between languages gradually narrows. High-performing models (e.g., DeepSeek-V3.2 and Qwen3-235B-A22B-Thinking-2507) leverage stronger cross-lingual semantic representation and deep reasoning abilities, which partially mitigate the impact of language differences, enabling more stable and balanced performance in multilingual long-context tasks. This trend also suggests that future LLMs are likely to further reduce language-induced performance gaps, achieving genuine cross-lingual consistency and universality.

\begin{figure*}[!t]
\centering
\includegraphics[width=0.88\textwidth]{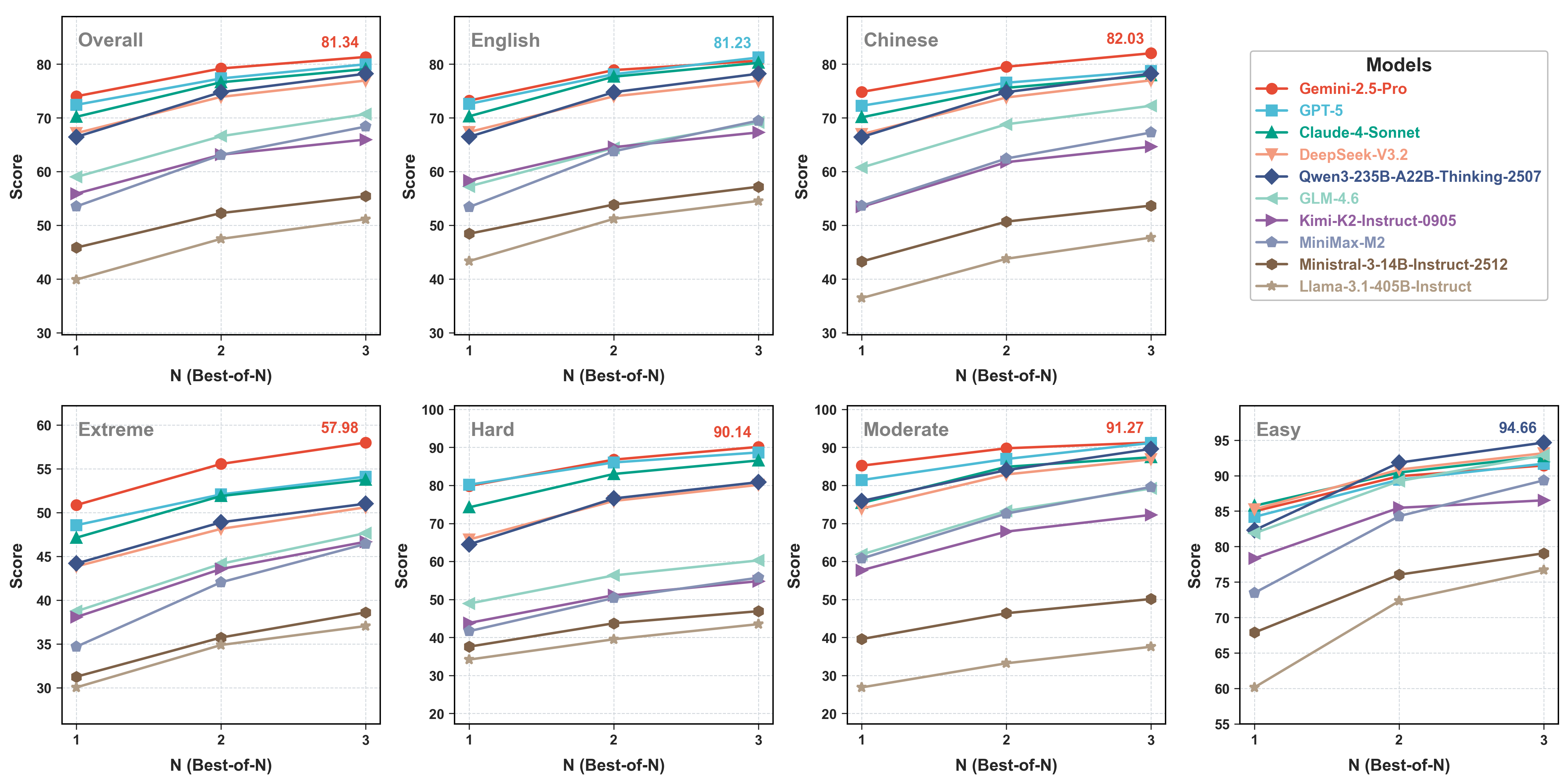}
\caption{Trends in Best-of-N metrics.}
\label{fig:bon_line}
\end{figure*}

\begin{figure*}[!h]
\centering
\includegraphics[width=0.88\textwidth]{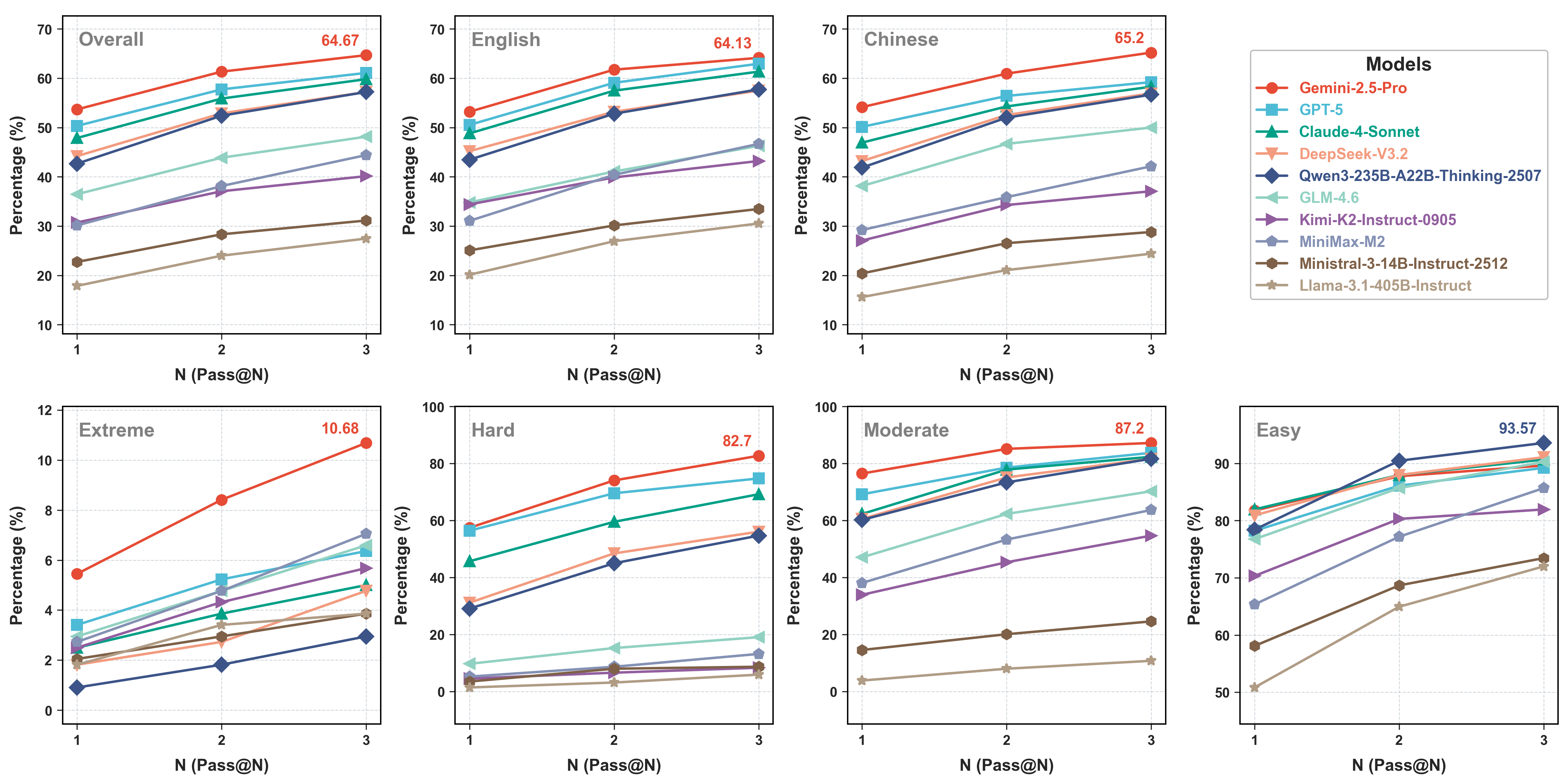}
\caption{Trends in Pass@N metrics.}
\label{fig:passn_line}
\end{figure*}

\textbf{(4) Extreme Difficulty Reveals the True Gap in Long-Context Capabilities.} The performance gap between open-source and closed-source models is minimal on Easy samples. For example, GPT-5 achieves 85.23, while DeepSeek-V3.2 achieves 85.02. However, the gap widens dramatically on the Extreme samples. For instance, Gemini-2.5-Pro scores 50.77, GPT-5 scores 48.37, DeepSeek-V3.2 scores 44.27, and Qwen3-235B-A22B-Thinking-2507 scores 43.39. Furthermore, the performance gains brought by ``thinking'' exhibit significant diminishing returns across tasks of different difficulty levels: the gains on Easy samples are much larger than those on Extreme samples. For example, after enabling thinking, Claude-4-Sonnet's score rises from 68.42 to 83.78 on Easy samples (+15.36), but only from 42.92 to 47.05 on Extreme samples (+4.13). Similarly, Gemini-2.5-Flash improves from 66.55 to 79.82 on Easy samples (+13.27), but only from 44.26 to 47.39 on Extreme samples (+3.13). These results indicate that the Extreme samples in LongBench Pro not only test a single capability of the models but also evaluate their combined abilities in long-context memory, integration, and reasoning. Current models still have considerable room for improvement on tasks of extreme difficulty.

\textbf{(5) The ``Thinking'' Paradigm Becomes a Key Breakthrough for Long-Context Performance.} Almost all models benefit from ``thinking''. For example, Gemini-2.5-Flash improves its score from 55.92 to 67.41 when thinking is enabled; Qwen3-235B-A22B-Instruct-2507 increases its score from 52.51 to 63.77 after performing thinking. Notably, Qwen3-4B with thinking enabled (40.82) even surpasses the non-thinking performance of Qwen3-32B (40.28), bridging the gap between different model sizes. These results indicate that long-context tasks involve cross-paragraph connections, which simple non-thinking modes easily miss, thereby limiting performance. Enabling the model to perform thinking becomes key to improving information retrieval and aggregation in long-context scenarios.

\textbf{(6) The Gap Between ``Native Thinking" and ``Prompted Thinking".} Not all models benefit from ``thinking." Thinking and mixed-thinking models internalize the thinking process, and the incorporation of thinking yields significant performance improvements for these models. For example, Claude-4-Sonnet achieves a gain of 13.80 (56.07 → 69.87); DeepSeek-V3.2 achieves a gain of 16.15 (51.67 → 67.82); and Qwen3-235B-A22B-Thinking-2507, compared with Qwen3-235B-A22B-Instruct-2507 with thinking enabled, still improves by 3.20 (63.77 → 66.97). Compared with them, traditional instruct models obtain very limited gains even when they are forced to think, and some smaller models even exhibit performance degradation caused by thinking. For instance, Llama-3.1-405B-Instruct yields only a 0.59 improvement (40.07 → 40.66); Gemma-3-12B-It shows a 0.24 decrease (32.16 → 31.92); and Llama-3.1-8B-Instruct suffers a 1.03 decrease (21.09 → 20.06). These results demonstrate that models without thinking training may fail to effectively leverage test-time compute. ``Thinking" is not merely a form of prompt engineering, but a fundamental post-training paradigm shift. Compared to prompted thinking, native thinking ability is the key to improving long-context performance.

\textbf{(7) Mixed-Thinking Models Exhibit Pareto Optimality.} Mixed-thinking models achieve Pareto-optimal performance between instruct models that cannot perform deep reasoning and thinking models that cannot respond quickly. They maintain efficient and robust baseline capability when thinking is disabled, and can approach or even surpass thinking models when thinking is enabled. For example, Gemini-2.5-Flash approaches the performance of Gemini-2.5-Pro in thinking mode, and DeepSeek-V3.2 significantly outperforms DeepSeek-R1 in thinking mode. This phenomenon indicates that mixed-thinking, which dynamically chooses between fast output and deep reasoning based on user needs, is highly likely to become the most competitive paradigm for future long-context models.

\begin{figure*}[!t]
\centering
\includegraphics[width=0.85\linewidth]{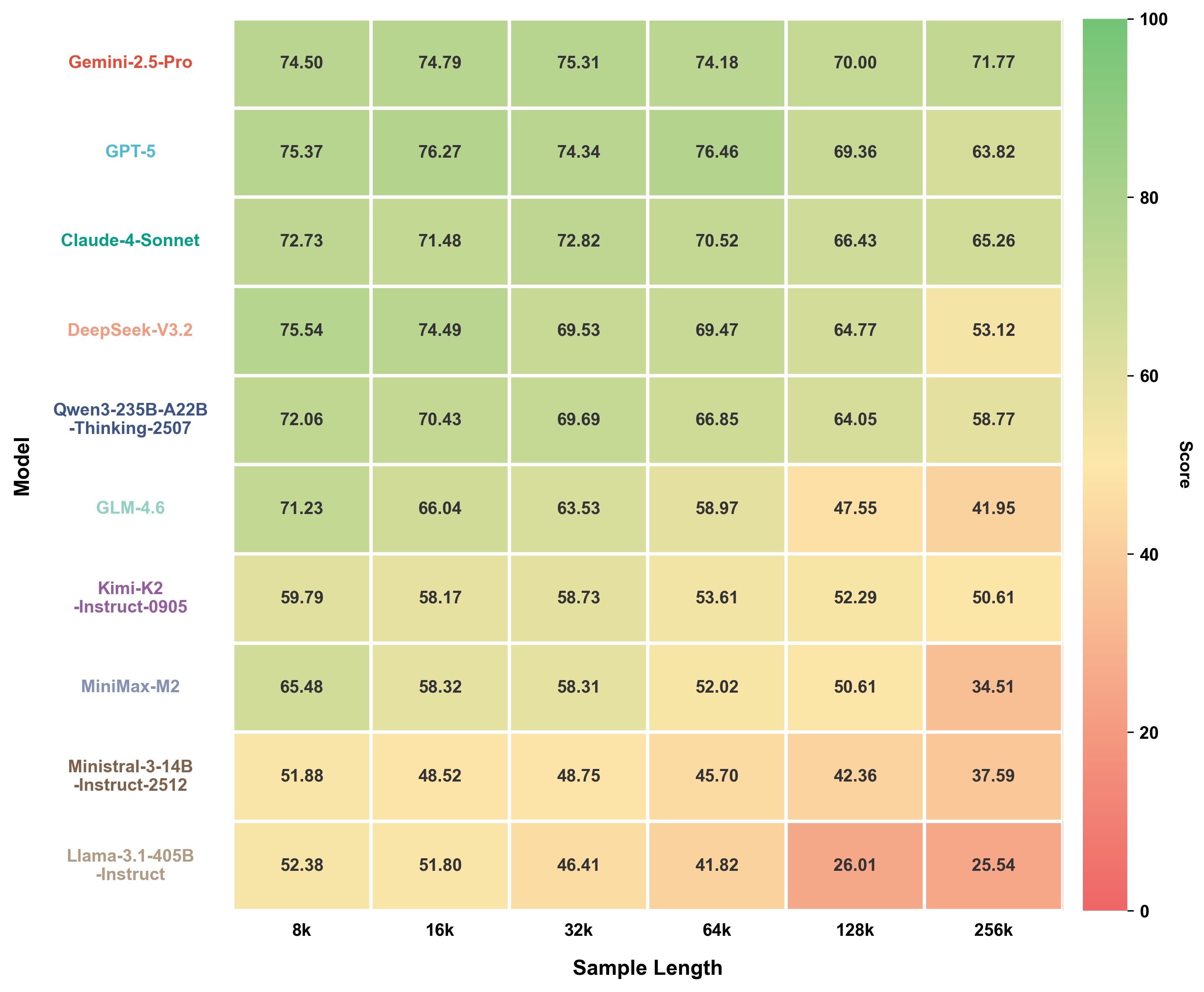}
\caption{Performance across different sample lengths.}
\label{fig:length_heatmap}
\end{figure*}

\begin{figure*}[h!]
\centering
\includegraphics[width=0.8\textwidth]{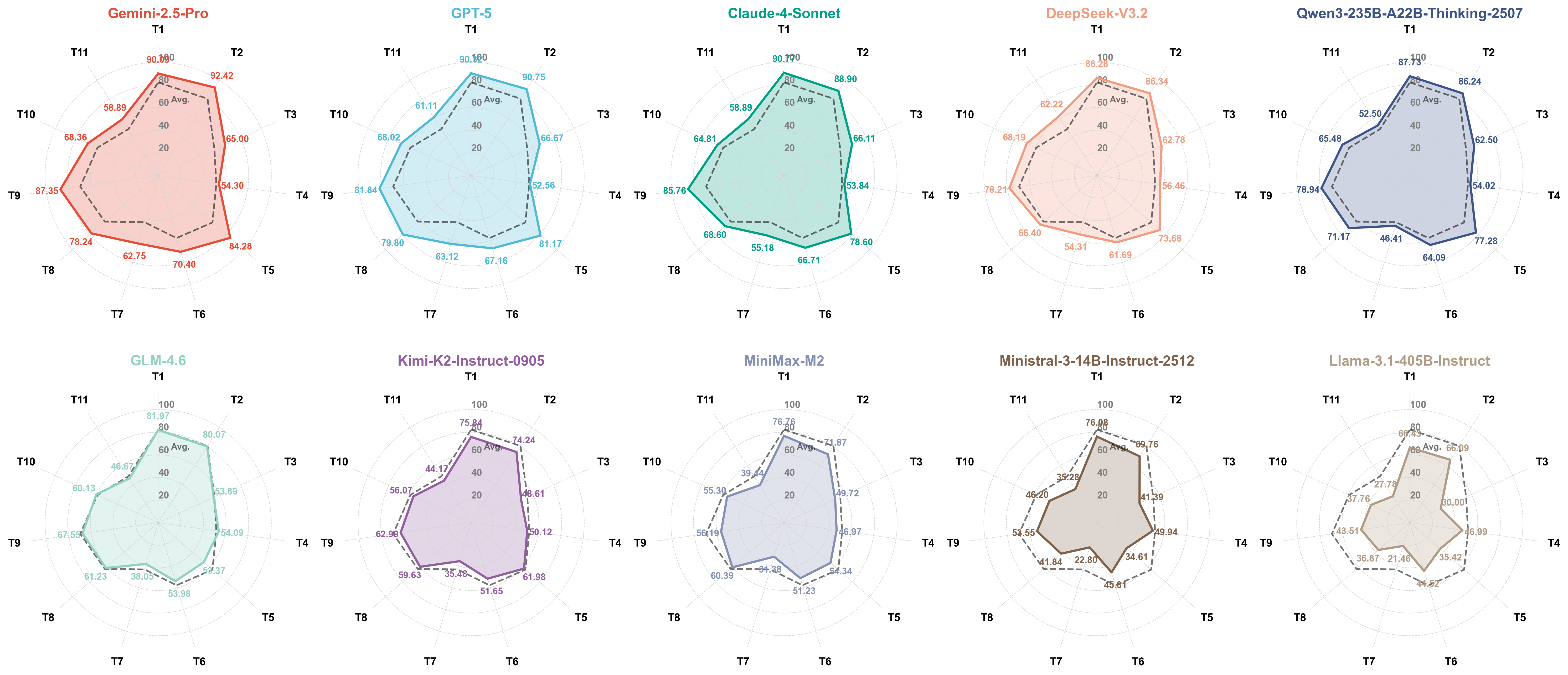}
\caption{Performance across different tasks.}
\label{fig:task_radar}
\end{figure*}

\subsection{Upper-Bound Performance}

To filter out metric deviations caused by generation instability, we report the trend of Best-of-N metrics for mainstream LLMs, as shown in Figure~\ref{fig:bon_line}. All models show a clear monotonic increase. Gemini-2.5-Pro and GPT-5 exhibit strong stability, as their single-shot performance is already high and the marginal gains from additional inference converge quickly. In contrast, models such as Qwen3-235B-A22B-Thinking-2507 exhibit markedly high potential: increasing N significantly corrects the bias in their initial reasoning, enabling them to make a leap toward the top tier.

LongBench Pro adopts a fine-grained scoring mechanism, granting partial credit for partially correct answers. This mechanism enables a more nuanced differentiation of performance across models; however, it fails to fully reflect the intrinsic difficulty of LongBench Pro. Therefore, we further report the trend of Pass@N metrics for mainstream LLMs on LongBench Pro, as shown in Figure~\ref{fig:passn_line}.

The above upper-bound evaluation substantiates the effectiveness and soundness of LongBench Pro. Under N = 3, LongBench Pro still maintains the following properties:

(1) Discriminability: There remains a clear performance gap across model tiers, indicating that LongBench Pro assesses deep long-context understanding rather than surface-level tricks that can be compensated for by probabilistic guessing.

(2) Difficulty: Even the strongest model, Gemini-2.5-Pro, achieves only a Pass@3 of 10.68 on the Extreme samples. After excluding factors related to model instability, the benchmark still exhibits a substantial headroom, sufficient to support the evaluation of more capable models.

\subsection{Comparison Across Length Dimension}

Figure~\ref{fig:length_heatmap} presents the performance levels of mainstream LLMs across different sample lengths. The results show that most models exhibit a declining trend in performance as sample length increases. For these models, sample length remains a significant factor affecting long-context performance. However, Gemini-2.5-Pro breaks this pattern, demonstrating remarkable length insensitivity: its score at 256k (71.77) is very close to its score at 8k (74.50). This phenomenon indicates that, within the range of 256k in length, for current state-of-the-art long-context models, merely increasing sample length to stress-test model performance has reached a point of saturation. The current bottleneck in long-context performance does not lie in the model's ability to ``read'' 256k tokens, but in its capacity to handle long-range dependencies and complex logical relationships. The focus of long-context evaluation shifts from ``how much can it read'' to ``how deeply can it comprehend,'' making the enhancement of models' deep comprehension ability in long contexts a major ongoing challenge.

\subsection{Comparison Across Task Dimension}

Figure~\ref{fig:task_radar} shows the performance levels of mainstream LLMs across different tasks. We make the following observations:

\textbf{(1) There is a significant gap between retrieval ability and aggregation ability.} Although most models demonstrate very high proficiency in basic information retrieval (T1) and sequence reconstruction (T2) (with average scores above 80), their performance drops sharply on semantic aggregation (T6, average score 57.72), which requires complex information integration. This contrast indicates that current models, although capable of precisely performing ``needle-in-a-haystack'' localizations, still face significant challenges when it comes to semantically aggregating and integrating dispersed information across long contexts.

\textbf{(2) There is an imbalance in forward and backward inference between evidence and outcomes.} Most models perform relatively well on evidence retrieval (T5, average score 63.47), but their performance is comparatively lower on question answering (T3) and summarization (T4), tasks that require deriving results from detailed document information (average scores below 55). This indicates that current models are more robust in backward alignment from outcomes to evidence, but forward generation from evidence to outcomes is more susceptible to document complexity and long-context effects.

\textbf{(3) Logical reasoning and consistency maintenance constitute current high-level bottlenecks.} Most models perform moderately to poorly on logical reasoning (T8–T10, average scores around 60), whereas some models enhanced with reasoning training perform exceptionally well on these tasks, highlighting the significant effect of targeted training on enhancing task-specific capabilities. In contrast, models generally score low on consistency maintenance (T7 and T11, average score below 49), exposing inherent limitations in sustaining global states over very long sequences.

\begin{figure*}[t!]
\centering
\includegraphics[width=0.8\textwidth]{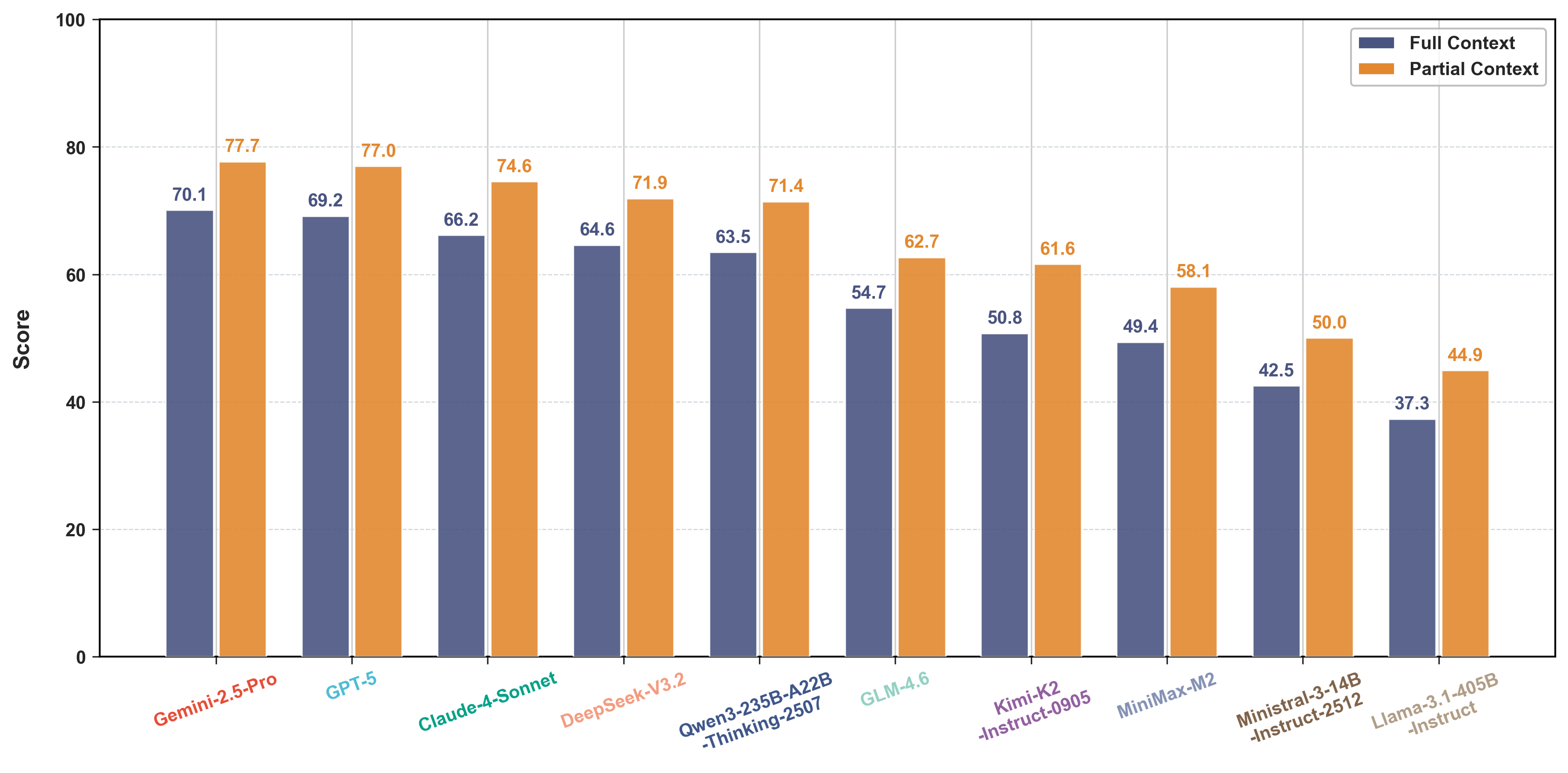}
\caption{Performance across different context requirements.}
\label{fig:context_requirement_bar}
\end{figure*}

\begin{figure*}[h!]
\centering
\includegraphics[width=1\textwidth]{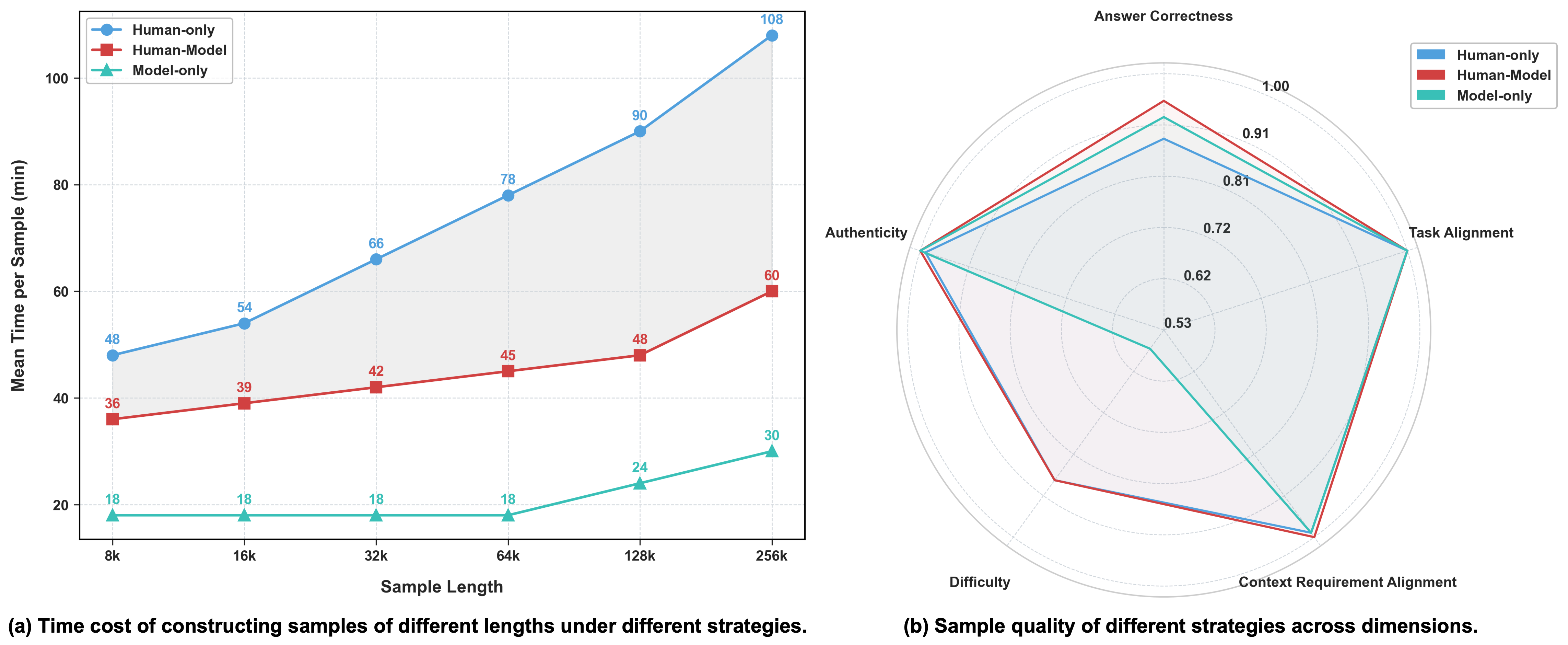}
\caption{Comparison of sample construction strategies.}
\label{fig:strategy_comparison}
\end{figure*}

\subsection{Comparison Across Context Requirement Dimension}

Figure~\ref{fig:context_requirement_bar} presents the performance differences of mainstream LLMs under varying context requirements. We observe a clear and widespread performance stratification: all models perform substantially better on Partial tasks, which emphasize localization and retrieval, than on Full tasks, which require integration and reasoning. Specifically, when questions shift from relying on a single local segment to requiring the integration of multiple segments across the entire document, the models exhibit a performance drop of 7.32 to 10.84 points. This result indicates that although current models demonstrate relatively mature capabilities in accurately retrieving local information from long contexts, they still show notable limitations in associating dispersed information across segments and performing holistic reasoning.

\subsection{Comparison Across Construction Strategies}
\label{sec:construction_strategies}

We uniformly sample 50 documents based on secondary task categories, lengths, and languages, and construct samples using three different strategies to evaluate the effectiveness of our strategy. The compared sample-construction strategies include human-only, model-only, and human-model collaboration (ours). Figure ~\ref{fig:strategy_comparison} (a) presents the time cost of constructing samples of different lengths under these strategies. The time required by the human-only strategy increases exponentially with sample length, while the model-only strategy remains consistently low. The human-model collaborative strategy falls between the other two and increases slowly with sample length, highlighting the substantial efficiency gains enabled by model involvement in the sample-construction process.

We also evaluate the quality of samples constructed using three strategies. Specifically, we design a sample quality assessment framework comprising five dimensions. According to this framework, each sample is scored by three experts on three levels, with scores ranging from 0 to 1. The results show that samples constructed via the human-model collaborative strategy achieve the highest average quality score (0.9609 ± 0.0415), outperforming those constructed by the human-only strategy (0.9484 ± 0.0450) and the model-only strategy (0.8964 ± 0.0536). The Fleiss' Kappa among the three experts is 0.76, indicating high agreement. Figure~\ref{fig:strategy_comparison} (b) shows the quality of samples constructed by different strategies across various dimensions. Thanks to our clear and explicit definitions of task and context requirements, as well as fully authentic natural text, all three sample construction strategies achieve consistently high scores in task alignment, context requirement alignment, and authenticity. On the difficulty dimension, the model-only strategy exhibits a relatively low level due to the absence of human sample filtering. In terms of answer correctness, the human-only strategy, without model assistance, tends to have omissions in answer components, resulting in the lowest correctness. The model-only strategy, without human error correction, tends to produce erroneous hallucinated answers, leading to slightly lower correctness. The human-model collaboration strategy effectively compensates for the weaknesses of both strategies, achieving the highest answer correctness.

\section{Related Works}

\label{sec:related_works}

Long-context evaluation measures whether LLMs can reliably retrieve, integrate, and reason over evidence that is sparse and distributed across lengthy documents, and it must address confounders such as positional effects and the gap between \emph{advertised} context length and \emph{effective} reasoning length~\cite{lostinthemiddle}. Existing benchmarks span a spectrum from controlled probes to realistic, human-verified tasks: synthetic benchmarks such as RULER~\cite{RULER}, MRCR~\cite{mrcr}, and GSM-$\infty$~\cite{gsm-infinite} provide scalable diagnostics of usable context size, while long-document NLP suites (e.g., SCROLLS/ZeroSCROLLS~\cite{scrolls,zeroscrolls}) and standardized evaluation protocols (e.g., L-Eval and its length-adaptable extension Ada-L-Eval~\cite{leval,adaleval}) broaden task coverage and comparability. More realistic benchmark datasets further emphasize natural documents and deeper reasoning, including LongBench~\cite{longbench} and LongBench v2~\cite{longbenchv2}, as well as mixed synthetic/natural stress tests such as $\infty$Bench~\cite{inftyBench} and multilingual evaluation such as CLongEval~\cite{Clongeval}; methodology-focused work like HELMET~\cite{Helmet} argues for systematic designs and analyses to avoid over-optimistic conclusions. In contrast to prior benchmarks that are primarily synthetic probes, protocol suites, or narrower in language/task coverage, LongBench Pro is built on fully natural long documents with bilingual (EN/ZH) coverage and diverse tasks/metrics, and it supports fine-grained analysis via multi-dimensional categorization (context requirement, length, difficulty) enabled by a scalable human-model collaborative construction pipeline.

\section{Conclusion and Future Work}

In this work, we introduce LongBench Pro, a realistic and comprehensive bilingual benchmark for long-context evaluation. We evaluate 46 representative long-context large language models (LLMs) on LongBench Pro and provide analyses across task, length, context requirement, difficulty, and language settings.

However, as task length and complexity continue to grow, even human-model collaborative construction can face a tension between verification accuracy and production efficiency. We are exploring a recursive critique scheme (``Critique-of-Critique''), which shares some similar ideas with the meta-verification design in DeepSeekMath-V2~\cite{deepseekmathv2}, to recursively and progressively decompose verification into easier subproblems that are tractable for human annotators. We have achieved preliminary results in this direction and look forward to sharing more findings in the near future.

\bibliography{example_paper}
\bibliographystyle{icml2025}



\newpage

\appendix

\section{Task Definitions}
\label{appendix:task_definitions}

\textcolor{t1dark}{\textbf{T1 Retrieval \& Ranking}} \\
Retrieve content and rank most relevant first. \\
\textcolor{t1dark}{\textbf{T1.1 Global Cohesive Retrieval}} \\
Retrieve full text and reorganize. \\
\textbf{Context Requirement:} Full \\
\textbf{Metric:} NDCG@k \\
\textbf{Example:}
\vspace{-2mm}
\begin{examplebox}
Retrieve all reviews that received 50 or more votes, and output the reviewerIDs in ascending order of vote value. Output the ``\texttt{[Answer]}" identifier first, and then output reviewerIDs line by line, without any additional content. \\
\\
Output example: \\
\texttt{[Answer]} \\
A2AV7Q95QGPTO0 \\
A3NM0RAYSL6PA8 \\
A1C9C1QOQB94RT \\
A1B80MVU7ZODF9
\end{examplebox}

\noindent\textcolor{t1dark}{\textbf{T1.2 Key-Snippet Retrieval}} \\
Locate target fragment in specified paragraph. \\
\textbf{Context Requirement:} Partial \\
\textbf{Metric:} NDCG@k \\
\textbf{Example:}
\vspace{-2mm}
\begin{examplebox}
From the subset of articles written by ``Dr. Seuss" (for which an artificial ID field has been added): Retrieve articles in the category ``Life" and sort them by popularity from lowest to highest. Output the ``\texttt{[Answer]}" identifier first, and then output the sorted reviewerID line by line, without any additional content. \\
\\
Output example: \\
\texttt{[Answer]} \\
ID1 \\
ID2 \\
ID3 \\
ID4
\end{examplebox}

\noindent\textcolor{t2dark}{\textbf{T2 Sequencing \& Structure Reconstruction}} \\
Restore timeline or logical order. \\
\textcolor{t2dark}{\textbf{T2.1 Global Timeline Reconstruction}} \\
Sort unordered events in the whole text. \\
\textbf{Context Requirement:} Full \\
\textbf{Metric:} Pairwise Accuracy \\
\textbf{Example:}
\vspace{-2mm}
\begin{examplebox}
This article is divided into nine parts, each part is preceded by ``Part x", and the order is shuffled. Please sort these nine parts in the correct order of events. Output the ``\texttt{[Answer]}" identifier first, and then output correct sequence of part numbers line by line, without any additional content. \\
\\
Output example: \\
\texttt{[Answer]} \\
Part 5 \\
Part 1 \\
Part 9 \\
Part 3 \\
Part 7 \\
Part 2 \\
Part 4 \\
Part 8 \\
Part 6
\end{examplebox}

\noindent\textcolor{t2dark}{\textbf{T2.2 Local Causal Chain Sorting}} \\
Sort content in a specific paragraph. \\
\textbf{Context Requirement:} Partial \\
\textbf{Metric:} Pairwise Accuracy \\
\textbf{Example:}
\vspace{-2mm}
\begin{examplebox}
The ``SELECTED PARAGRAPH" of the article is out of order. Please reorder this paragraph according to the original text and options. Output the ``\texttt{[Answer]}" identifier first, and then output the sorted option letters line by line, without any additional content. \\
\\
Output example: \\
\texttt{[Answer]} \\
A \\
B \\
C \\
D \\
E 
\end{examplebox}

\noindent\textcolor{t3dark}{\textbf{T3 Evidence-Grounded QA}} \\
Answer fact/reasoning questions based on evidence. \\
\textcolor{t3dark}{\textbf{T3.1 Multi-Doc Integration QA}} \\
Use multi-hop information to answer questions. \\
\textbf{Context Requirement:} Full \\
\textbf{Metric:} Accuracy \\
\textbf{Example:}
\vspace{-2mm}
\begin{examplebox}
On their way to Egloshayle, from the conversation of the passengers on the coach, it can be inferred that the characteristics of Mr. Alan Torrington do not include the  following? Output the ``\texttt{[Answer]}" identifier first, and then output the answer option letter (A/B/C/D), without any additional content. \\
\\
A. He was rumored by some to be a reclusive person, with a habit of wandering near the rocks at night, and was even falsely labeled as a ``mysterious wizard." \\
B. His personality was in stark contrast to that of his brother Oscar Torrington, and he was considered to be a miser, with rumors that he hid his wealth in secret places. \\
C. He was in poor health, and his trip to Switzerland with his brother was aimed at improving his health, as he usually required someone to take care of his daily life. \\
D. He had once placed a light in the tower window to mislead ships into running aground in order to acquire their cargo, becoming a figure associated with the local history of ``wrecking." \\
\\
Output example: \\
\texttt{[Answer]} \\
C
\end{examplebox}

\noindent\textcolor{t3dark}{\textbf{T3.2 Single-Hop Fact QA}} \\
Answer questions based on local paragraphs. \\
\textbf{Context Requirement:} Partial \\
\textbf{Metric:} Accuracy \\
\textbf{Example:}
\vspace{-2mm}
\begin{examplebox}
What are all the topic categories listed in the paper ``Delta Activities: A Representation for Finetuned Large Language Models" (arXiv: 2509.04442)? Output the ``\texttt{[Answer]}" identifier first, and then output the answer option letter (A/B/C/D), without any additional content. \\
\\
A. Machine Learning (cs.LG);  Computation and Language (cs.CL);  Artificial Intelligence (cs.AI) \\
B. Artificial Intelligence (cs.AI);  Computation and Language (cs.CL);  Information Retrieval (cs.IR) \\
C. Machine Learning (cs.LG);  Artificial Intelligence (cs.AI);  Computation and Language (cs.CL);  Information Retrieval (cs.IR) \\
D. Machine Learning (cs.LG);  Artificial Intelligence (cs.AI);  Information Retrieval (cs.IR);  Sound (cs.SD) \\
\\
Output example: \\
\texttt{[Answer]} \\
A
\end{examplebox}

\noindent\textcolor{t4dark}{\textbf{T4 Summarization \& Synthesis}} \\
Generate abstract summary under given constraints. \\
\textcolor{t4dark}{\textbf{T4.1 Global-Coverage Constrained Summary}} \\
Generate summary of full text. \\
\textbf{Context Requirement:} Full \\
\textbf{Metric:} 0.5*max(SemSim)+0.5*max(ROUGE-L) \\
\textbf{Example:}
\vspace{-2mm}
\begin{examplebox}
The above are multiple chapters of a novel, and based on the given text, summarize and generalize the entire content, with a requirement of no more than 100 words. Output the ``\texttt{[Answer]}" identifier first, and then output the summary, without any additional content. \\
\\
Output example: \\
\texttt{[Answer]} \\
your summary
\end{examplebox}

\noindent\textcolor{t4dark}{\textbf{T4.2 Query-Focused Summary}} \\
Generate summary of specific subtopic. \\
\textbf{Context Requirement:} Partial \\
\textbf{Metric:} 0.5*max(SemSim)+0.5*max(ROUGE-L) \\
\textbf{Example:}
\vspace{-2mm}
\begin{examplebox}
Summarize the plot of the absurd victory celebration only in the first section ($\le$ 200 words). Output the ``\texttt{[Answer]}" identifier first, and then output the summary, without any additional content. \\
\\
Output example: \\
\texttt{[Answer]} \\
your summary
\end{examplebox}

\noindent\textcolor{t5dark}{\textbf{T5 Attribution \& Citation Alignment}} \\
Bind correct sources to generated text. \\
\textcolor{t5dark}{\textbf{T5.1 Full-Sentence Citation Alignment}} \\
Citation alignment for all sentences. \\
\textbf{Context Requirement:} Full \\
\textbf{Metric:} F1 \\
\textbf{Example:}
\vspace{-2mm}
\begin{examplebox}
You will be provided with a summary sentence. Your task is to identify the original Part number(s) from the provided text that fully support this sentence. Output the ``\texttt{[Answer]}" identifier first, and then output the cited content (format as ``Part xx") line by line, without any additional content. \\
\\
Generated Summary: \\
After taking a leave of absence post the 1984 Macintosh launch (to recharge and avoid supervisor Bob Belleville), Andy Hertzfeld clashed with Jobs first. Over a denied Mac team bonus (which Jobs initially blamed on Belleville. Before relenting, leaving Hertzfeld upset) and later when Jobs dismissed his. Concerns about the Mac software team's low morale and Burrell's frustration. As out of touch, insisting the team was doing well. \\
\\
Output example: \\
\texttt{[Answer]} \\
Part 1 \\
Part 12 \\
Part 15
\end{examplebox}

\noindent\textcolor{t5dark}{\textbf{T5.2 Key-Statement Citation Alignment}} \\
Citation alignment for specified sentences. \\
\textbf{Context Requirement:} Partial \\
\textbf{Metric:} F1 \\
\textbf{Example:}
\vspace{-2mm}
\begin{examplebox}
You will see a generated 3-sentence summary. Identify and cite the original source location for Sentence 2 only, using the paragraph identifiers S1–S120 from the source text. Output the ``\texttt{[Answer]}" identifier first, and then output the paragraph identifier(s) line by line, without any additional content. \\
\\
Generated summary: \\
Sentence 1: During Danielle Mitterrand's factory tour, she pressed on labor conditions—overtime and vacation—while Jobs, increasingly annoyed, emphasizedautomation and just-in-time processes. \\
Sentence 2: Jobs believed that Sculley did not understand business and was not suitable to manage the company, and designed a plan to remove him. \\
Sentence 3: Driving back toward Cupertino, Jobs was ticketed for speeding near 100 mph and, unfazed, accelerated again—evidence, Rossmann said, that Jobs believed normal rules didn't apply to him. \\
\\
Output example: \\
\texttt{[Answer]} \\
S1 \\
S2
\end{examplebox}

\noindent\textcolor{t6dark}{\textbf{T6 Aggregation \& Clustering}} \\
Cluster and output statistics/examples/sort. \\
\textcolor{t6dark}{\textbf{T6.1 Large-Scale Document Clustering}} \\
Return all category proportions. \\
\textbf{Context Requirement:} Full \\
\textbf{Metric:} SubEM \\
\textbf{Example:}
\vspace{-2mm}
\begin{examplebox}
Cluster the 8 documents (IDs: A–H) by research methodology into exactly three clusters: Randomized trial/protocol; Quantitative observational; Qualitative. Output the ``\texttt{[Answer]}" identifier first, and then output the three clusters in the format ``ClusterName Proportion\% (rounded to two decimal places)" line by line, without any additional content. \\
\\
Output example: \\
\texttt{[Answer]} \\
Randomized trial/protocol 25.00\% \\
Quantitative observational 25.00\% \\
Qualitative 50.00\%
\end{examplebox}

\noindent\textcolor{t6dark}{\textbf{T6.2 Targeted Subset Cluster Identification}} \\
Return query category instances. \\
\textbf{Context Requirement:} Partial \\
\textbf{Metric:} F1 \\
\textbf{Example:}
\vspace{-2mm}
\begin{examplebox}
Find the first three software reviews in the Software Reviews category. Output the ``\texttt{[Answer]}" identifier first, and then output the answer reviewer IDs that meet the conditions line by line, without any additional content. \\
\\
Output example: \\
\texttt{[Answer]} \\
A22V1MD93T2FW9 \\
ACJT8MUCOLRFO \\
A38NELQT98S4H8
\end{examplebox}

\noindent\textcolor{t6dark}{\textbf{T6.3 Global Frequency Analysis}} \\
Count and sort global word frequency. \\
\textbf{Context Requirement:} Full \\
\textbf{Metric:} Pairwise Accuracy \\
\textbf{Example:}
\vspace{-2mm}
\begin{examplebox}
Sort the given terms in descending order by the number of times they appear in the above text. Output the ``\texttt{[Answer]}" identifier first, and then output the sorted terms line by line, without any additional content. \\
\\
Given terms: would, this, that, went, have \\
\\
Output example: \\
\texttt{[Answer]} \\
would \\
this \\
that \\
went \\
have
\end{examplebox}

\noindent\textcolor{t7dark}{\textbf{T7 Consistency \& Compliance Checking}} \\
Detect and locate contradictions/violations. \\
\textcolor{t7dark}{\textbf{T7.1 Global Conflict \& Inconsistency Localization}} \\
Locate contradictory segments in the full text. \\
\textbf{Context Requirement:} Full \\
\textbf{Metric:} F1 \\
\textbf{Example:}
\vspace{-2mm}
\begin{examplebox}
Compare the inconsistent articles between clauses A and B in two documents. Output the ``\texttt{[Answer]}" identifier first, and then output all inconsistent articles line by line in the format ``A[articleID] B[articleID]", without any additional content. \\
\\
Output example: \\
\texttt{[Answer]} \\
A1 B2 \\
A2 B3
\end{examplebox}

\noindent\textcolor{t7dark}{\textbf{T7.2 Targeted Rule or Condition Violation Detection}} \\
Locate content that violates specific rules. \\
\textbf{Context Requirement:} Partial \\
\textbf{Metric:} F1 \\
\textbf{Example:}
\vspace{-2mm}
\begin{examplebox}
Read and list which terms of Enzuzo, Inc. Privacy Policy were violated in the first article. Output the ``\texttt{[Answer]}" identifier first, and then output all the violated term IDs line by line, without any additional content. \\
\\
Output example: \\
\texttt{[Answer]} \\
5 \\
6
\end{examplebox}

\noindent\textcolor{t7dark}{\textbf{T7.3 Comprehensive Error \& Anomaly Sweep}} \\
Locate spelling errors in the full text. \\
\textbf{Context Requirement:} Full \\
\textbf{Metric:} F1 \\
\textbf{Example:}
\vspace{-2mm}
\begin{examplebox}
Find and list all misspelled English words in a given complete text file. Output the ``\texttt{[Answer]}" identifier first, and then output all misspelled English words line by line, without any additional content. \\
\\
Output example: \\
\texttt{[Answer]} \\
beleive \\
freind \\
tommorow
\end{examplebox}

\noindent\textcolor{t8dark}{\textbf{T8 Structured \& Numeric Reasoning}} \\
Numerical calculations in structured text. \\
\textcolor{t8dark}{\textbf{T8.1 Structured Multi-Source Consistency Verification}} \\
Numerical computation in multi-source. \\
\textbf{Context Requirement:} Full \\
\textbf{Metric:} SubEM \\
\textbf{Example:}
\vspace{-2mm}
\begin{examplebox}
Above are two tables. Please use the following formula to determine if there are any issues after merging the two tables (i.e., whether the sales amount calculated using the formula for Table 1 is consistent with the sales amount in Table 2). If there are no issues, return ``No Error". If there are any issues, return the corresponding Invoice IDs. Output the ``\texttt{[Answer]}" identifier first, and then output ``No Error" or output corresponding Invoice IDs line by line, without any additional content. \\
\\
Formula: Sales Amount = Unit Price × Quantity + 5\% Tax \\
\\
Output example: \\
\texttt{[Answer]} \\
No Error/226 31 3081
\end{examplebox}

\noindent\textcolor{t8dark}{\textbf{T8.2 Single-Source Targeted Aggregation}} \\
Query computation in single-source. \\
\textbf{Context Requirement:} Partial \\
\textbf{Metric:} SubEM \\
\textbf{Example:}
\vspace{-2mm}
\begin{examplebox}
According to the table ``Share repurchases for the fourth quarter of fiscal year 2023", calculate the percentage increase in the ``Total Number of Shares Purchased as Part of Publicly Announced Plans or Programs" for June compared to April and May. Output the ``\texttt{[Answer]}" identifier first, and then output percentages (with one decimal place retained) line by line, without any additional content. \\
\\
Output example: \\
\texttt{[Answer]} \\
5.5\% \\
-2.2\%
\end{examplebox}

\noindent\textcolor{t8dark}{\textbf{T8.3 Long-Context Procedural State Tracking}} \\
Track entity state evolution. \\
\textbf{Context Requirement:} Full \\
\textbf{Metric:} F1 \\
\textbf{Example:}
\vspace{-2mm}
\begin{examplebox}
Find and list all misspelled English words in a given complete text file. Output the ``\texttt{[Answer]}" identifier first, and then output all misspelled English words line by line, without any additional content. \\
\\
Output example: \\
\texttt{[Answer]} \\
beleive \\
freind \\
tommorow
\end{examplebox}

\noindent\textcolor{t9dark}{\textbf{T9 Version \& Code Diff Analysis}} \\
Compare changes in different text/code versions. \\
\textcolor{t9dark}{\textbf{T9.1 Dependency-Aware Multi-Version Impact Analysis}} \\
Track dependency changes across versions. \\
\textbf{Context Requirement:} Full \\
\textbf{Metric:} F1 \\
\textbf{Example:}
\vspace{-2mm}
\begin{examplebox}
Based on the two provided API documents for hadoop-core (version 0.20.0 and 0.21.0), identify all methods within the ``org.apache.hadoop.fs.FileSystem" class that were not deprecated in version 0.20.0 but became deprecated in version 0.21.0. Use the method signature as the identifier in the format ``MethodName(ParameterTypes)". If a method has no parameters, use ``MethodName()". For multiple parameters, use a comma-separated list of fully qualified types. Output the ``\texttt{[Answer]}" identifier first, and then output the identifiers line by line, without any additional content. \\
\\
Output example: \\
\texttt{[Answer]} \\
MethodA(java.lang.String) \\
MethodB() \\
MethodC(int, java.lang.String)
\end{examplebox}

\noindent\textcolor{t9dark}{\textbf{T9.2 Localized Interface Change Detection}} \\
Detect local version differences. \\
\textbf{Context Requirement:} Partial \\
\textbf{Metric:} F1 \\
\textbf{Example:}
\vspace{-2mm}
\begin{examplebox}
Identify the fields within the ``Lesson" resource model that were either renamed, removed due to architectural changes, or refactored into a more complex structure when evolving from V1.0 to V2.0. List these fields using the format ``Lesson Model: [V2.0 Field Name]" for renamed/refactored fields and ``Lesson Model: [V1.0 Field Name]" for fields that were entirely removed. Output the ``\texttt{[Answer]}" identifier first, and then output fields line by line, without any additional content. \\
\\
Output example: \\
\texttt{[Answer]} \\
Lesson Model: key \\
Lesson Model: seq\_length
\end{examplebox}

\noindent\textcolor{t10dark}{\textbf{T10 Rule Induction \& In-Context Learning}} \\
Summarize rules and make decisions on new samples. \\
\textcolor{t10dark}{\textbf{T10.1 Large-Scale In-Context Rule Induction}} \\
Induce rules from the global context. \\
\textbf{Context Requirement:} Full \\
\textbf{Metric:} SubEM \\
\textbf{Example:}
\vspace{-2mm}
\begin{examplebox}
You are a script formatter. Based on the full range of script formatting conventions demonstrated in the provided text, reformat the following raw script snippet. Pay close attention to scene headings, character names, dialogue parenthetical actions, and transitions. Output the ``\texttt{[Answer]}" identifier first, and then output the fully formatted script snippet, without any additional content. \\
\\
Raw Snippet: \\
interior, lab - day \\
Dr. Banner (looking at monitor): the readings are stable now.
Tony Stark: That's what I like to hear. \\
(Tony smiles) \\
cut to: \\
exterior, city rooftop - night \\
A figure (hooded) stands looking at the skyline. \\
\\
Output example: \\
\texttt{[Answer]} \\
your answer
\end{examplebox}

\noindent\textcolor{t10dark}{\textbf{T10.2 Targeted Example-Based Rule Induction}} \\
Induce rules from the targeted examples. \\
\textbf{Context Requirement:} Partial \\
\textbf{Metric:} SubEM \\
\textbf{Example:}
\vspace{-2mm}
\begin{examplebox}
Answer the following questions based on the content of the case in Part 1 of the provided document: UKCo is a private limited company specializing in cross-border private equity investments. In 2022, UKCo signed a Joint Investment Agreement with three EU investment institutions (A, B, C), agreeing to jointly  establish a European infrastructure cooperation plan to invest in EU new energy infrastructure projects. In 2024, the European Infrastructure  Cooperation Program fell into a debt crisis due to project defaults, owing 8  million euros to supplier Company D (a German company). At the same time, A, B, and C found that UKCo had misappropriated funds from the European Infrastructure Cooperation Scheme in its management, so they applied to the  English Commercial Court: Should the English court support A, B, and C's  application for compulsory liquidation of the European Infrastructure  Cooperation Scheme as a compulsory liquidation? Output the ``\texttt{[Answer]}" identifier first, and then output answer. The answer must be a fixed value (``support/don't support"), and without any additional content. \\
\\
Output example: \\
\texttt{[Answer]} \\
support
\end{examplebox}

\noindent\textcolor{t11dark}{\textbf{T11 Dialogue Memory \& Long-Horizon Tracking}} \\
Track and respond to dialogue history. \\
\textcolor{t11dark}{\textbf{T11.1 Long-Range Entity \& Commitment Tracking}} \\
Track entity states across the global context. \\
\textbf{Context Requirement:} Full \\
\textbf{Metric:} Accuracy \\
\textbf{Example:}
\vspace{-2mm}
\begin{examplebox}
Based on the conversation between the user and the model above, which of the following options best fits the content in the text? Output the ``\texttt{[Answer]}" identifier first, and then output A/B/C/D/E, without any additional content. \\
\\
A. About the book ``Ship of Theseus", Based on the conversation, here is a simple and insightful summary of the book ``S." / ``Ship of Theseus": ``S." is a unique, single level reading experience that casts the reader as a detective.It is a book within a book, containing. \\
B. About the book ``Piranesi", Based on the conversation, here is a summary of the discussion about Susanna Clarke's novel ``Piranesi": A signature style of the author, Susanna Clarke, She uses authentic academic citations, which lends the story a remarkable sense of authenticity and depth while also concealing clues. \\
C. About the movie ``Lobster", Based on the conversation, here is a summary of the analysis of the movie ``The Lobster": The discussion analyzes Yorgos Lanthimos's 2015 film ``The Lobster," identifying its core premise-a society where single people must find a partner within a month or become an animal-as a satirical critique of modern society's pressure to be in a relationship. \\
D. About ``Dark City", The user and model discuss the 1999 Alex Proyas film ``Dark City", agreeing it is an underrated sci-fi classic often overshadowed by ``The Matrix". They identify its core premise: a city perpetually trapped in night where alien ``Strangers" experiment on humans by altering their memories and physical reality to understand the human soul, which they themselves lack. \\
E. ``The Ship of Theseus", The Joy of Collaboration: A central theme is that the best way to enjoy these complex works is through collaboration. The model encourages the user to solve puzzles oneself, suggesting they can divide tasks and that debating theories will lead to deeper insights. This is supported by real-world examples of fan communities and is also applied to finding hidden details (like changing tie patterns) in the film ``Predestination" (``Former Destination"). \\
\\
Output example: \\
\texttt{[Answer]} \\
B
\end{examplebox}

\noindent\textcolor{t11dark}{\textbf{T11.2 Short-Range Reference Resolution \& State Query}} \\
Resolve references and states in local context. \\
\textbf{Context Requirement:} Partial \\
\textbf{Metric:} Accuracy \\
\textbf{Example:}
\vspace{-2mm}
\begin{examplebox}
In the scene in the elegant dining room where Cobb and Arthur are pitching their services to Saito (pages 2-4), what specific reason does Arthur give for why a person's thoughts become vulnerable to theft in the dream state? Output the ``\texttt{[Answer]}" identifier first, and then output the answer option letter (A/B/C/D), without any additional content. \\
\\
A. Because a person can be trained to reveal their own secrets. \\
B. Because killing someone in a dream will not wake them up. \\
C. Because a person's conscious defenses are lowered. \\
D. Because an idea, once it takes hold, is impossible to eradicate. \\
\\
Output example: \\
\texttt{[Answer]} \\
B
\end{examplebox}

\section{Annotation Guidelines}
\label{appendix:annotation_guidelines}

\subsection{Sample-Generation Prompt}

\begin{guidelinebox}
\textbf{Role Definition} \\
You are a professional question-design specialist who: \\
•	possesses solid capabilities in textual deconstruction and deep reading; \\
•	has extensive cross-domain experience in exam item construction; \\
•	demonstrates accurate command of the following domains: legal and regulatory texts, literature and creative writing, academic and professional materials, consulting and media content, history and culture, as well as general-interest texts; \\
•	is familiar with each domain's technical terminology, logical systems, textual features, and core assessment points. \\
\\
\textbf{Task Objective} \\
Given an input \textbf{long context} and its \textbf{task category (primary task + secondary task)}, you create \textbf{three questions} that strictly fulfill the task requirements (including \textbf{answers}, \textbf{design rationale}, and detailed \textbf{solution process}). \\
\\
\textbf{Task Procedure} \\
\textbf{Step 1: Language Identification} \\
Determine the language of the input long text (e.g., Chinese/English). All subsequent content (questions, answers, explanations) must \textbf{fully use the same language} as the long context. \\
\textbf{Step 2: Question Construction} \\
\textbf{2.1 Number of Questions} \\
Based on the long context and the secondary task type, extract deep assessment points and generate \textbf{three questions} that satisfy the secondary task specifications. \\
\textbf{2.2 Construction Requirements} \\
•	All questions must strictly satisfy the evaluation goals of the primary task, the required I/O format, and the definition of the secondary task. \\
•	Question phrasing must follow the secondary task's example patterns. \\
•	Answer formats must match the task examples. \\
•	The three questions must exhibit clear differentiation, avoiding repeated formats or duplicate assessment points. \\
\textbf{2.3 Sample Requirements} \\
\textit{(1) Identifier Rules (ensuring answer uniqueness)} \\
To avoid ambiguity caused by overlapping textual expressions, each question must apply identifiers: \\
\ding{192} Natural identifiers (preferred). Use naturally occurring, unique elements in the document as identifiers, such as titles, section names, or explicit entities. \\
\ding{193} Constructed identifiers. If natural identifiers are insufficient, create identifiers for document segments, e.g., DocumentX ParagraphX IDX D11. Requirements: \\
•	Use only letters and numbers; do not include symbols (such as ``- , \_ ."). \\
•	Explicitly explain the numbering rule in the question. \\
•	Require respondents to answer using the identifiers. \\
\textit{(2) Requirements for Multiple-Choice Questions (if applicable)} \\
•	Must include distractors based on the document. \\
•	At least four options. \\
•	The correct option must be verifiable in the text. \\
•	Distractors must be reasonable and sufficiently plausible. \\
\textit{(3) Reference Answer (for summarization tasks only)} \\
•	Three results are required for semantic similarity evaluation; must be accurate and cover key points. \\
\textit{(4) Question-Design Rationale (required for every question) }\\
You must explain: \\
•	how the question satisfies the task type; \\
•	which document information the question draws upon; \\
•	why the answer is unique and verifiable. \\
\textit{(5) Solution and Evidence (required for every question)} \\
You must provide: \\
•	the complete, correct answer; \\
•	detailed solution steps; \\
•	clear citation of all textual evidence, with one-to-one correspondence. \\
\textit{(6) Unified Output Format} \\
•	All questions must require line-by-line answers to support automated evaluation. \\
\textbf{2.4 Task Types} \\
\textit{(1) Context Requirement Dimension} \\
•	Full: The context appears in fully scrambled order. A complete reading of the entire text is necessary to retrieve all content required for the question; missing any segment leads to an inaccurate answer. \\
•	Partial: Since the relevant segments are contiguous and ordered, one only needs to locate the target segment; the rest is not essential. \\
\textit{(2) Specific Task Type} \\
\textbf{\textcolor{blue!70!black}{\{Primary\_Task\_Definition\}}} \\
\textbf{\textcolor{blue!70!black}{\{Secondary\_Task\_Definition, Context\_Requ \\ -irement, I/O\_Specification, and Examples\}}} \\
\\
\textbf{Step 3: Self-Check} \\
After generating the three questions, you must conduct a rigorous self-check (no need to output the self-check content): \\
•	Whether the questions fully comply with the task type and secondary task requirements; \\
•	Whether the answers are accurate, verifiable, and free from speculation; \\
•	Whether the questions contain no ambiguity, avoid repeated assessment points, and exhibit sufficient difficulty; \\
•	Whether the domain-specific expression is correct (e.g., legal or classical-text domains); \\
•	Whether the language of the questions and answers matches the long context. \\
If any criterion fails → You must return to Step 2 and regenerate the questions. \\
\\
\textbf{Input} \\
\texttt{[Long Context]}: \textbf{\textcolor{blue!70!black}{\{Long\_Context\}}} \\
\texttt{[Primary Task]}: \textbf{\textcolor{blue!70!black}{\{Primary\_Task\}}} \\
\texttt{[Secondary Task]}: \textbf{\textcolor{blue!70!black}{\{Secondary\_Task\}}} \\
\\
\textbf{Output Constraints} \\
1.	Output three questions, and the question types must match the secondary task requirements. \\
2.	The language of the questions, answers, and explanations must match the long text. \\
3.	The output format must strictly contain: the questions, the answers, the question-design rationale, and the solution process. No additional content may be included.
\end{guidelinebox}

\subsection{Sample Verification Criteria}

\begin{guidelinebox}
\textbf{Task Objective} \\
This task systematically and formally verifies question–answer samples initially generated by large models. The goal is to ensure that each sample meets the requirements \textbf{in task authenticity, reasoning correctness, and challenge level.} \\
\\
\textbf{Sample Generation Process} \\
To balance sample authenticity with human labor costs, we adopt a human-model collaborative approach to generate test samples: \\
\textit{1.	Multi-model Generation} \\
For each long document, annotators generate three questions and answers using the following state-of-the-art models with specific prompts. Each model output includes: the question, the answer, the design rationale, and the solving process. \\
o	Gemini-2.5-Pro \\
o	GPT-5 \\
o	Claude-Sonnet-4 \\
o	DeepSeek-V3.2 \\
o	Qwen3-235B-A22B-Thinking-2507 \\
\textit{2.	Human Critique} \\
Annotators critically review the model outputs and decide whether to adopt, modify, or discard each sample. \\
\textit{3.	Secondary Review} \\
Expert perform a final audit of generated samples; any sample failing the review returns for further revision. \\
\\
\textbf{Verification Tasks Overview} \\
Annotators complete the following three assessments for each model-generated sample: \\
\textbf{Task A: Question Compliance with Task Type and Context Requirement} \\
Based on the model's ``design rationale," annotators evaluate whether the question: \\
•	Aligns with the task definition and context requirement. \\
•	Relies on document content only, without using external information. \\
•	Avoids subjective judgment or unverifiable inference. \\
•	Remains on-topic and does not include unsupported facts. \\
\textbf{Task B: Answer Correctness} \\
Based on the model's ``solving process," annotators evaluate whether: \\
•	The reasoning steps are grounded in the document. \\
•	The logic is internally consistent. \\
•	The final answer is verifiable from the document. \\
•	The answer does not contain hallucinations (fabricated information). \\
\textbf{Task C: Question Challenge Level} \\
Annotators input the question into the five models listed above and assess: \\
•	At least one model answers incorrectly → the sample is considered challenging. \\
•	If all five models answer correctly → the sample is not challenging. \\
\\
\textbf{Verification Procedure} \\
\textbf{\textit{Step 1}}: Read the model output, including the question, answer, design rationale, and solving process. Annotators fully understand the intended assessment points of the question. \\
\textbf{\textit{Step 2}}: Execute Task A - evaluate question compliance: \\
o	If the question does not match the task type or diverges from the document → discard the sample. \\
o	If minor edits can fix the issue → modify and proceed to Step 3. \\
o	If fully compliant → proceed to Step 3. \\
\textbf{\textit{Step 3}}: Execute Task B - evaluate answer correctness by checking against the original document and reasoning process: \\
o	If the answer is incorrect or the reasoning contains hallucinations/jumps → discard the sample. \\
o	If minor edits can fix the issue → modify and proceed to Step 4. \\
o	If fully correct → proceed to Step 4. \\
\textbf{\textit{Step 4}}: Execute Task C - verify challenge level by testing the question with five models: \\
o	If all models answer correctly → the question is not challenging; discard unless the question is highly valuable. \\
o	If at least one model answers incorrectly → the question is challenging; proceed to Step 5. \\
\textbf{\textit{Step 5}}: Select the best sample among all candidates: \\
o	If multiple samples pass → select the optimal one (most aligned with definitions and highest difficulty) and add it to the sample set. \\
o	If no sample passes → proceed to the next document. \\
\textbf{\textit{Step 6}}: Submit for expert review: \\
o	Annotators submit verified samples to long-context expert. \\
o	If experts reject → revise according to expert feedback and resubmit until approval. \\
\\
\textbf{Precautions} \\
\textbf{Prohibited Actions} \\
•	Do not fabricate information not present in the document. \\
•	Do not treat model hallucinations as facts. \\
•	Do not use subjective questions. \\
•	Do not generate unverifiable answers. \\
•	Avoid overly simple questions. \\
\textbf{Permitted/Encouraged Actions} \\
•	Retain questions that require deep reasoning, cross-paragraph inference, or implicit clues. \\
•	Refine model-generated questions to improve quality. \\
•	Use the model reasoning chain to assist verification, but final judgment must be human-led. \\
\\
\textbf{Sample Quality Checklist} \\
\begin{tabularx}{\linewidth}{|X|c|}
\hline
\textbf{Item} & \textbf{Check} \\
\hline
Task type: Matches design & $\square$ \\
Document reliance: Context only & $\square$ \\
Clarity: Question is clear & $\square$ \\
Answer verifiable: Checkable & $\square$ \\
Reasoning: Grounded logic & $\square$ \\
Challenge: At least one model fails & $\square$ \\
Readability: Clear format & $\square$ \\
\hline
\end{tabularx}
\end{guidelinebox}

\subsection{Sample Rewriting Criteria}

\begin{guidelinebox}
\textbf{Task Objective} \\
Standardize the prompts of different question types to a uniform format to improve the accuracy and success rate of subsequent automatic evaluation. Each question requires two prompts: \\
•	\textbf{Non-Thinking Prompt} (does not output the thinking process) \\
•	\textbf{Thinking Prompt} (outputs the thinking process first, then the answer) \\
Both types of prompts share the same structure, with only minor differences in Output Requirement and Output Example.\\
\\
\textbf{Unified Structure (Mandatory)} \\
Each prompt follows the three sections below: \\
\textbf{\textit{\ding{192} First Section}}: Task Description + Output Requirement (Mandatory) \\
•	Task Description: Explain what the question asks to do. \\
•	Output Requirement: Write differently for the Non-Thinking or Thinking version (see below). \\
\textbf{\textit{\ding{193} Second Section}}: Supplementary Content (Optional) \\
If the question contains additional materials (e.g., abstract, full text, options, label list), include them in this section. Omit this section entirely if no supplementary content exists. \\
\textbf{\textit{\ding{194} Third Section}}: Output Example (Mandatory) \\
The output example must: \\
•	Fully comply with the required answer format. \\
•	Differ from the real answer content. \\
•	Contain the ``\texttt{[Answer]}" identifier. \\
\\
\textbf{Differences Between the Two Prompt Types (Core Part)} \\
The following content is used to rewrite Output Requirement and Output Example. The rest (Task Description, Supplementary Content) remains unchanged. \\
\textbf{1. Non-Thinking Prompt Specification} \\
\textit{Output Requirement (Non-Thinking Version) — Fixed Template, Cannot Change} \\
\verb|```|  \\
Output the ``\texttt{[Answer]}" identifier first, and then output \{elements\} line by line, without any additional content. \\
\verb|```| \\
Here, elements are provided by the annotator according to the question, e.g., sorted numbers, option letters, titles, years, or custom items. The format of each element must be clear, e.g., ``Number Paragraph", ``A", ``Item1 Item2". \\
\textit{Output Example (Non-Thinking Version) — Fixed Format} \\
\verb|```| \\
Output example: \\
\texttt{[Answer]} \\
\{Specific element 1\} \\
\{Specific element 2\} \\
\verb|```| \\
\textbf{2. Thinking Prompt Specification} \\
\textit{Output Requirement (Thinking Version) — Fixed Template, Cannot Change} \\
\verb|```| \\
Think step by step. After your thinking process, output the ``\texttt{[Answer]}" identifier, and then output \{elements\} line by line. \\
\verb|```| \\
The elements are also specified according to the question. \\
\textit{Output Example (Thinking Version) — Fixed Format} \\
\verb|```| \\
Output example: \\
<Your thinking process> \\
\texttt{[Answer]} \\
\{Specific element 1\} \\
\{Specific element 2\} \\
\verb|```| \\
\\
\textbf{Prompt Rewriting Procedure (Mandatory for Annotators)} \\
The following steps are used to rewrite the original question into the standardized format. \\
\textbf{\textit{Step 1}}: Break Down Original Question Content \\
Extract four parts from the original sample: \\
•	Task Description \\
•	Output Requirement (use fixed template for Non-Thinking / Thinking) \\
•	Supplementary Content (if any) \\
•	Output Example (write according to requirements) \\
\textbf{\textit{Step 2}}: Construct Non-Thinking Prompt \\
\textbf{Template} \\
\verb|```| \\
{Task Description} Output the ``\texttt{[Answer]}" identifier first, and then output \{elements\} line by line, without any additional content. \\
\\
{Supplementary Content (optional)} \\
\\
Output example: \\
\texttt{[Answer]} \\
\{Specific element 1\} \\
\{Specific element 2\} \\
\verb|```| \\
\textbf{\textit{Step 3}}: Construct Thinking Prompt \\
\textbf{Template} \\
\verb|```| \\
{Task Description} Think step by step. After your thinking process, output the ``\texttt{[Answer]}" identifier, and then output \{elements\} line by line. \\
\\
{Supplementary Content (optional)} \\
\\
Output example: \\
<Your thinking process> \\
\texttt{[Answer]} \\
\{Specific element 1\} \\
\{Specific element 2\} \\
\verb|```| \\
\textbf{\textit{Step 4}}: Final Review \\
Check each item: \\
•	The three-section structure is complete. \\
•	Both Non-Thinking and Thinking formats use the fixed templates. \\
•	Example format fully matches the answer format and differs from the real answer. \\
•	``\texttt{[Answer]}" identifier exists and is correct. \\
•	No mixing of Chinese and English punctuation.
\end{guidelinebox}

\subsection{Answer Review Criteria}

\begin{guidelinebox}
\textbf{Task Objective} \\
To ensure the accuracy and reliability of all sample answers, we perform consistent, standardized, and high-quality verification of long-context test samples. The verification process follows these principles: \\
•	Human annotators are skilled at judging whether each component of an answer is correct → responsible for \textbf{precision}. \\
•	Models can generate a diverse set of possible answer components → responsible for \textbf{recall}. \\
Therefore, during verification, we combine human judgment with model predictions to ensure data completeness. \\
\\
\textbf{Verification Procedure} \\
\textbf{\textit{Step 1}}: Generate model predictions
For each sample, we pre-generate a list of predicted answers from five state-of-the-art models. \\
\textbf{\textit{Step 2}}: Initial annotator verification (two annotators) \\
•	\textit{Step A}: Check the correctness of each component in the current answer (ensures precision). \\
•	\textit{Step B}: Check for any missing reasonable components, referring to model predictions if necessary (ensures recall). \\
\textbf{\textit{Step 3}}: Result determination \\
•	If both annotators identify no issues → mark the sample as problem-free and directly include it in the benchmark. \\
•	If any annotator identifies a potential issue → send the sample to a long-context expert for final judgment. \\
•	The long-context expert determines, based on the issue type, whether the sample needs reconstruction. \\
\\
\textbf{Three Types of Potential Issues} \\
During verification, three types of issues require identification: \\
\textbf{\ding{192} Document Issues} \\
The provided document content is incomplete, lacking the necessary information to derive the current answer. \\
\textbf{\ding{193} Question Issues} \\
The question description is unclear, incomplete, or ambiguous, preventing a reasonable derivation of the current answer. \\
\textbf{\ding{194} Answer Issues} \\
The document and question are fine, but the existing answer is incorrect or incomplete. \\
\\
\textbf{Annotator Operational Guidelines} \\
\textbf{1.	Independent judgment} \\
Both annotators must complete verification independently to avoid human consistency bias. \\
\textbf{2.	Ensure precision} \\
Evaluate the correctness of each ``component" in the answer to guarantee fine-grained assessment. \\
\textbf{3.	Ensure recall} \\
Check for any missing components; the model prediction list serves as a reference. \\
o	If a reasonable component appears in model predictions but is not covered in the current answer → mark as missing. \\
o	If model predictions are unreasonable → do not consider it an issue. \\
\textbf{4.	Flag any doubts} \\
Any uncertainty or potential error must be flagged for review by the long-context expert. \\
\\
\textbf{Long-Context Experts Guidelines} \\
The long-context expert is responsible for: \\
1.	Making final judgments on disputed samples. \\
2.	Determining whether a sample requires reconstruction. \\
3.	If reconstruction is needed, specifying the exact reason and corresponding step (modify document, rewrite question, or revise answer).
\end{guidelinebox}

\subsection{Sample Quality Evaluation Criteria}

\begin{guidelinebox}
\textbf{Task Objective} \\
Each long-context question is scored along the following five evaluation dimensions. Each dimension allows scores of 0, 0.5, or 1. \\
\\
\textbf{Evaluation Dimensions and Criteria} \\
\textbf{1. Task Alignment} \\
\textbf{\textit{Definition:}} Measures whether the question aligns with the predefined task objectives, i.e., whether it matches the intended task design (25 secondary tasks in total). \\
\textbf{\textit{Scoring Criteria:}} \\
•	1: Fully aligns with the task objective; the question format matches the task type closely. \\
•	0.5: Largely aligns, but the task intention is unclear or slightly deviates. \\
•	0: Clearly deviates from the task objective or uses an incorrect task type. \\
\textbf{2. Context Requirement Alignment} \\
\textit{\textbf{Definition:}} Evaluates whether the question's dependence on document information matches the intended context requirement level. For example, Full questions require the entire document, while Partial questions require only local paragraphs. \\
\textbf{\textit{Scoring Criteria:}} \\
•	1: The context dependency fully matches the intended level. \\
•	0.5: The dependency level is generally reasonable but slightly deviates. \\
•	0: The dependency level is incorrect (e.g., a Partial question requires the entire document). \\
\textbf{3. Difficulty} \\
\textbf{\textit{Definition:}} Measures the challenge and discriminative power of the question, estimated based on the accuracy of five models answering the question. \\
\textbf{\textit{Scoring Criteria:}} \\
•	1: Accuracy is between 0\%–20\%; at most one model answers correctly. \\
•	0.5: Accuracy is between 40\%–60\%; two to three models answer correctly. \\
•	0: Accuracy is between 80\%–100\%; four to five models answer correctly. \\
\textbf{4. Authenticity} \\
\textbf{\textit{Definition:}} Measures whether the question reflects real user information needs and inquiry style, i.e., whether the question is natural and meaningful in practical scenarios. \\
\textbf{\textit{Scoring Criteria:}} \\
•	1: Natural tone, fluent expression, reasonable question with authentic motivation. \\
•	0.5: Generally natural but slightly templated or artificial. \\
•	0: Clearly synthetic or inconsistent with real-world question style. \\
\textbf{5. Answer Correctness} \\
\textbf{\textit{Definition:}} Measures whether the reference answer is accurate, consistent with the document, and reasonably verifiable. \\
\textbf{\textit{Scoring Criteria:}} \\
•	1: Answer fully matches the document; facts are correct. \\
•	0.5: Answer is mostly correct with minor deviations or imprecise wording. \\
•	0: Answer is clearly incorrect or inconsistent with the document. \\
\\
\textbf{Scoring Procedure} \\
1.	\textbf{Independent Scoring}: Each sample is independently scored by three experts along the five dimensions. \\
2.	\textbf{Score Level}s: Each dimension allows only 0, 0.5, or 1. \\
3.	\textbf{Result Aggregation}: The final score for each dimension is the average of the three experts' scores. \\
4.	\textbf{Notes}: \\
o	Scoring strictly follows the criteria without subjective assumptions. \\
o	For ambiguous cases, the context and task type may be considered, but consistency must be maintained. \\
o	All scores require reasonable explanations for subsequent review.
\end{guidelinebox}

\section{Annotator Statistics and Compensation}

\begin{figure*}[h!]
\centering
\includegraphics[width=1\textwidth]{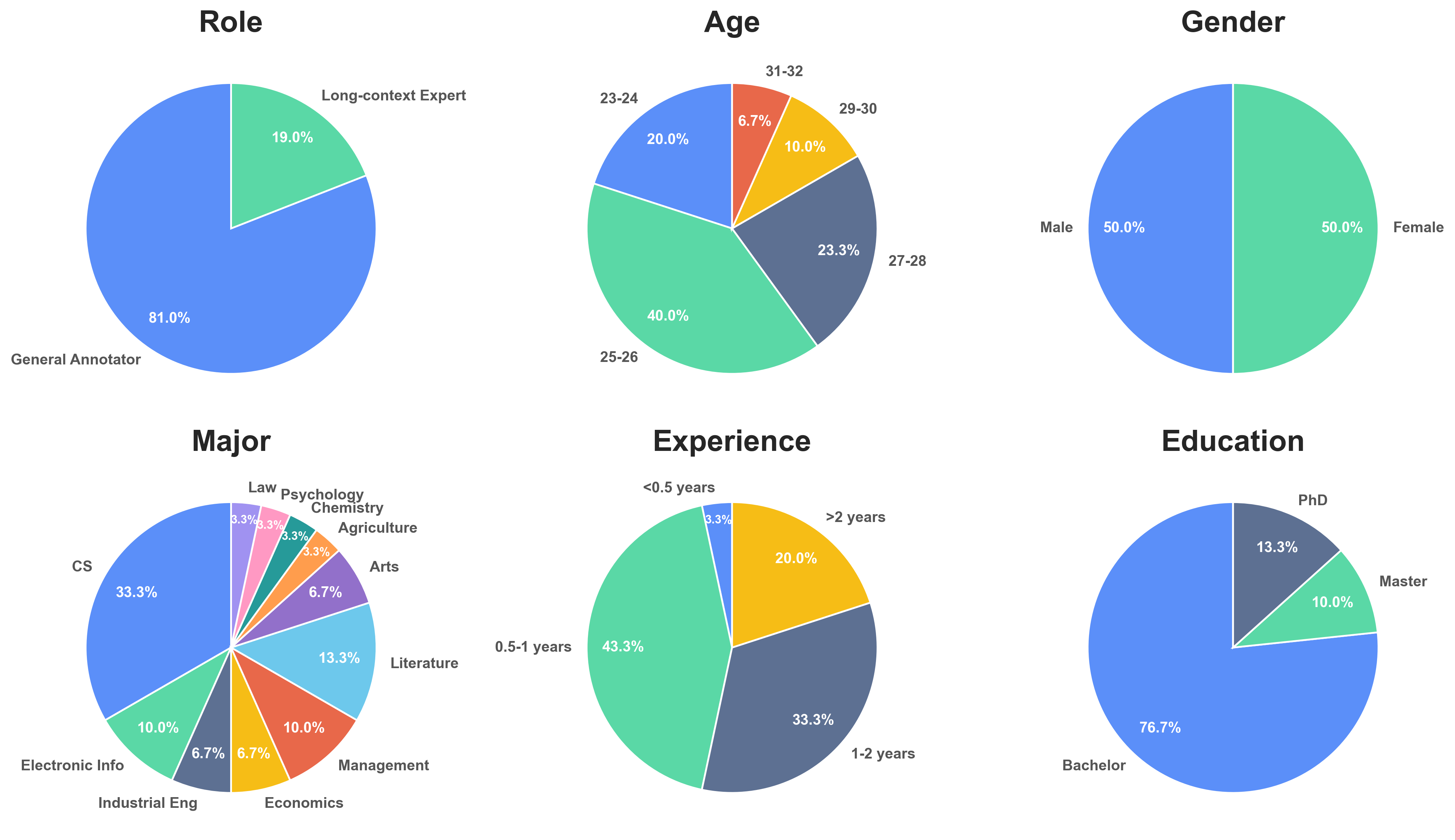}
\caption{Distribution of annotators' role, age, gender, major, experience, and education.}
\label{fig:annotator_stat_pie}
\end{figure*}

The construction of LongBench Pro involves a total of 63 annotators, with the statistical distribution shown in Figure~\ref{fig:annotator_stat_pie}. The 63 annotators are divided into 51 general annotators and 12 long-context experts (all long-text experts have at least one year of annotation experience and receive a two-month specialized training in long-context annotation). The ages of the annotators mainly range from 23 to 32 years, the gender ratio is balanced, and their major backgrounds are diverse. In addition, more than half of the annotators have over one year of annotation experience. Most annotators hold a bachelor's degree, and approximately 25\% possess a graduate degree. Annotators are compensated at a rate of 50 RMB per hour.

\section{Inference Parameter Settings}
\label{appendix:inference_parameter_settings}

Table~\ref{tab:inference_parameter_settings} presents the detailed inference parameter settings of different models, with the parameters taken from the open-source documentation of each model.

\section{Truncation Length Setting}
\label{appendix:truncation_length_setting}

The actual output length of some thinking models is often greater than the default 8k setting. For example, in MiniMax-M2, 538 out of 4,500 inferences have lengths far exceeding 8k. Setting the output length to 8k would result in a large number of evaluation failures. Therefore, to reserve sufficient thinking space, we set the truncation length of DeepSeek-V3.2, GLM-4.6, and MiniMax-M2 to 120k, ensuring that the output length can be set to 32k.

In addition, there is a significant discrepancy between the effective context length and the claimed context length for some models. Table~\ref{tab:glm46_truncation_lengths} presents a comparison of the non-thinking scores of GLM-4.6 (which claims a context length of 198k) under truncation lengths of 190k and 120k. When the truncation length is set to 190k, for 256k samples (whose length approaches 190k), the model outputs become unstable, leading to a sharp drop in metrics. This indicates that GLM-4.6's effective context length significantly deviates from its claimed context length. This is also one of the reasons why we set its truncation length to 120k.

\begin{table}[h!]
\centering
\resizebox{\columnwidth}{!}{
\begin{tabular}{c|cccccc}
\toprule
\textbf{Truncation} & \multicolumn{6}{c}{\textbf{Sample Length}} \\
\textbf{Length} & \textbf{8k} & \textbf{16k} & \textbf{32k} & \textbf{64k} & \textbf{128k} & \textbf{256k} \\
\midrule
190k & 53.74 & 50.76 & 51.93 & 45.60 & 37.73 & \textbf{\textcolor{red!70!black}{2.55}} \\
120k & 53.98 & 49.88 & 52.26 & 46.18 & 38.68 & 34.14 \\
\bottomrule
\end{tabular}}
\caption{Non-thinking scores of GLM-4.6 on samples of different lengths under varying truncation lengths.}
\label{tab:glm46_truncation_lengths}
\end{table}

\begin{table*}[ht!]
\centering
\resizebox{\textwidth}{!}{%
\begin{tabular}{cl | c | cc | c | cc | cc}
\toprule
\multicolumn{2}{c|}{\multirow{2}{*}{\textbf{Model}}} &
\textbf{Model} &
\multicolumn{2}{c|}{\textbf{Context Length}} &
\textbf{Truncation} &
\multicolumn{2}{c|}{\textbf{Output Length}} &
\multicolumn{2}{c}{\textbf{Temperature}} \\

\multicolumn{2}{c|}{} & 
\textbf{Type} & 
\textbf{Input} &
\textbf{Output} &
\textbf{Length} &
\textbf{Non-Thk.} &
\textbf{Thk.} &
\textbf{Non-Thk.} &
\textbf{Thk.} \\

\midrule

\multirow{5}{*}{\includegraphics[height=1.5em]{gemini.png}}

& Gemini-2.5-Pro      & Thinking & 1M   & 64k & 1M   & -  & 32k & -  & 1.0 \\
& Gemini-2.5-Flash    & Mixed    & 1M   & 64k & 1M   & 1k & 32k & 1.0 & 1.0 \\
& Gemma-3-27B-It      & Instruct & \multicolumn{2}{c|}{128k} & 120k & 1k & 8k  & 1.0 & 1.0 \\
& Gemma-3-12B-It      & Instruct & \multicolumn{2}{c|}{128k} & 120k & 1k & 8k  & 1.0 & 1.0 \\
& Gemma-3-4B-It       & Instruct & \multicolumn{2}{c|}{128k} & 120k & 1k & 8k  & 1.0 & 1.0 \\

\midrule

\multirow{4}{*}{\includegraphics[height=1.5em]{openai.png}} 

& GPT-5             & Thinking & 272k & 128k & 272k & -  & 32k & -  & 1.0 \\
& GPT-4o            & Instruct & \multicolumn{2}{c|}{128k} & 120k & 1k & 8k  & 1.0 & 1.0 \\
& GPT-OSS-120B      & Thinking & \multicolumn{2}{c|}{128k} & 120k & -  & 8k  & - & 1.0 \\
& GPT-OSS-20B       & Thinking & \multicolumn{2}{c|}{128k} & 120k & -  & 8k  & - & 1.0 \\

\midrule

\multirow{2}{*}{\includegraphics[height=1.5em]{claude.png}} 

& Claude-4-Sonnet     & Mixed    & 1M   & 64k & 1M   & 1k & 32k & 1.0 & 1.0 \\
& Claude-3.7-Sonnet   & Mixed    & 200k & 128k & 200k & 1k & 32k & 1.0 & 1.0 \\

\midrule

\multirow{5}{*}{\includegraphics[height=1.5em]{deepseek.png}} 

& DeepSeek-V3.2 \textbf{\textcolor{red!70!black}{*}}        & Mixed    & \multicolumn{2}{c|}{160k} & 120k & 1k  & 32k & 1.0 & 1.0 \\
& DeepSeek-V3.1       & Mixed    & \multicolumn{2}{c|}{128k} & 120k & 1k  & 8k  & 0.6 & 0.6 \\
& DeepSeek-R1-0528    & Thinking & \multicolumn{2}{c|}{128k} & 120k & -   & 8k  & -   & 0.6 \\
& DeepSeek-R1         & Thinking & \multicolumn{2}{c|}{128k} & 120k & -   & 8k  & -   & 0.6 \\
& DeepSeek-V3-0324    & Instruct & \multicolumn{2}{c|}{128k} & 120k & 1k  & 8k  & 0.3 & 0.3 \\

\midrule

\multirow{13}{*}{\includegraphics[height=1.5em]{qwen.png}} 

& Qwen3-235B-A22B-Thinking-2507 & Thinking  & \multicolumn{2}{c|}{256k} & 224k & -   & 32k & -   & 0.6 \\
& Qwen3-235B-A22B-Instruct-2507 & Instruct & \multicolumn{2}{c|}{256k} & 224k & 1k  & 32k & 0.7 & 0.7 \\
& Qwen3-Next-80B-A3B-Thinking    & Thinking  & \multicolumn{2}{c|}{256k} & 224k & -   & 32k & -   & 0.6 \\
& Qwen3-Next-80B-A3B-Instruct    & Instruct & \multicolumn{2}{c|}{256k} & 224k & 1k  & 32k & 0.7 & 0.7 \\
& Qwen3-30B-A3B-Thinking-2507    & Thinking  & \multicolumn{2}{c|}{256k} & 224k & -   & 32k & -   & 0.6 \\
& Qwen3-30B-A3B-Instruct-2507    & Instruct & \multicolumn{2}{c|}{256k} & 224k & 1k  & 32k & 0.7 & 0.7 \\
& Qwen3-4B-Thinking-2507         & Thinking  & \multicolumn{2}{c|}{256k} & 224k & -   & 32k & -   & 0.6 \\
& Qwen3-4B-Instruct-2507         & Instruct & \multicolumn{2}{c|}{256k} & 224k & 1k  & 32k & 0.7 & 0.7 \\
& Qwen3-32B                        & Mixed     & \multicolumn{2}{c|}{128k} & 120k & 1k  & 8k  & 0.7 & 0.6 \\
& Qwen3-14B                        & Mixed     & \multicolumn{2}{c|}{128k} & 120k & 1k  & 8k  & 0.7 & 0.6 \\
& Qwen3-8B                         & Mixed     & \multicolumn{2}{c|}{128k} & 120k & 1k  & 8k  & 0.7 & 0.6 \\
& Qwen3-4B                         & Mixed     & \multicolumn{2}{c|}{128k} & 120k & 1k  & 8k  & 0.7 & 0.6 \\
& Qwen2.5-72B-Instruct             & Instruct  & \multicolumn{2}{c|}{128k} & 120k & 1k  & 8k  & 0.7 & 0.7 \\

\midrule

\multirow{2}{*}{\raisebox{-0.5em}{\includegraphics[height=1.5em]{zai.png}} } 

& GLM-4.6 \textbf{\textcolor{red!70!black}{*} }    & Mixed    & \multicolumn{2}{c|}{198k} & 120k & 1k  & 32k & 1.0 & 1.0 \\
& GLM-4.5    & Mixed    & \multicolumn{2}{c|}{128k} & 120k & 1k  & 8k  & 1.0 & 1.0 \\

\midrule

\raisebox{-0.5em}{\includegraphics[height=1.5em]{kimi.png}}

& Kimi-K2-Instruct-0905 & Instruct & \multicolumn{2}{c|}{256k} & 224k & 1k & 32k & 0.6 & 0.6 \\

\midrule

\multirow{2}{*}{\includegraphics[height=1.5em]{minimax.png}} 

& MiniMax-M2 \textbf{\textcolor{red!70!black}{*}} & Thinking & \multicolumn{2}{c|}{192k} & 120k & -   & 32k & -   & 1.0 \\
& MiniMax-Text-01     & Instruct & \multicolumn{2}{c|}{4M}  & 1M   & 1k  & 32k & 1.0 & 1.0 \\

\midrule

\multirow{7}{*}{\includegraphics[height=1.5em]{mistral.png}} 

& Ministral-3-14B-Instruct-2512    & Instruct  & \multicolumn{2}{c|}{256k} & 224k & 1k  & 32k & 0.1 & 0.1 \\
& Ministral-3-8B-Instruct-2512     & Instruct  & \multicolumn{2}{c|}{256k} & 224k & 1k  & 32k & 0.1 & 0.1 \\
& Ministral-3-3B-Instruct-2512     & Instruct  & \multicolumn{2}{c|}{256k} & 224k & 1k  & 32k & 0.1 & 0.1 \\
& Magistral-Small-2509              & Thinking  & \multicolumn{2}{c|}{128k} & 120k & -   & 8k  & -   & 0.7 \\
& Mistral-Small-3.2-24B-Instruct-2506 & Instruct  & \multicolumn{2}{c|}{128k} & 120k & 1k  & 8k  & 0.15 & 0.15 \\
& Mistral-Large-Instruct-2411      & Instruct  & \multicolumn{2}{c|}{128k} & 120k & 1k  & 8k  & 1.0 & 1.0 \\
& Ministral-8B-Instruct-2410       & Instruct  & \multicolumn{2}{c|}{128k} & 120k & 1k  & 8k  & 1.0 & 1.0 \\

\midrule

\multirow{5}{*}{\includegraphics[height=1.5em]{meta.png}} 

& Llama-3.1-405B-Instruct & Instruct & \multicolumn{2}{c|}{128k} & 120k & 1k & 8k  & 0.6 & 0.6 \\
& Llama-3.3-70B-Instruct  & Instruct & \multicolumn{2}{c|}{128k} & 120k & 1k & 8k  & 0.6 & 0.6 \\
& Llama-3.1-70B-Instruct  & Instruct & \multicolumn{2}{c|}{128k} & 120k & 1k & 8k  & 0.6 & 0.6 \\
& Llama-3.1-8B-Instruct   & Instruct & \multicolumn{2}{c|}{128k} & 120k & 1k & 8k  & 0.6 & 0.6 \\
& Llama-3.2-3B-Instruct   & Instruct & \multicolumn{2}{c|}{128k} & 120k & 1k & 8k  & 0.6 & 0.6 \\

\bottomrule
\end{tabular}}
\caption{Detailed inference parameter settings. \textbf{Non-Thk.} denotes Non-Thinking, and \textbf{Thk.} denotes Thinking. \textbf{Length} is uniformly measured by the number of tokens. \textbf{\textcolor{red!70!black}{*}}: Although these models support longer context lengths, we set their truncation length uniformly to 120k and the thinking output length to 32k to enable more thorough thinking. Appendix~\ref{appendix:truncation_length_setting} provides specific notes regarding this part.}
\label{tab:inference_parameter_settings}
\end{table*}

\end{document}